\documentclass{article}

\usepackage[final]{neurips_2024}

\usepackage[utf8]{inputenc} 
\usepackage[T1]{fontenc}    
\usepackage{hyperref}       
\usepackage{url}            
\usepackage{booktabs}       
\usepackage{amsfonts}       
\usepackage{nicefrac}       
\usepackage{microtype}      
\usepackage{xcolor}         
\usepackage{multirow}
\usepackage{graphicx}

\usepackage[colorinlistoftodos,prependcaption,textsize=small]{todonotes}

\title{Different Horses for Different Courses: Comparing Bias Mitigation Algorithms in ML}

\author{%
    Prakhar Ganesh\thanks{Equal contribution} \\
    McGill University and Mila\\
    \texttt{prakhar.ganesh@mila.quebec} \\
    \And
    Usman Gohar$^*$ \\
    Iowa State University \\
    \texttt{ugohar@iastate.edu} \\
    \AND
    Lu Cheng \\
    University of Illinois Chicago \\
    \texttt{lucheng@uic.edu} \\
    \And
    Golnoosh Farnadi \\
    McGill University and Mila \\
    \texttt{farnadig@mila.quebec} \\
}

\begin{document}

\maketitle

\begin{abstract}
With fairness concerns gaining significant attention in Machine Learning (ML), several bias mitigation techniques have been proposed, often compared against each other to find the best method. These benchmarking efforts tend to use a common setup for evaluation under the assumption that providing a uniform environment ensures a fair comparison. However, bias mitigation techniques are sensitive to hyperparameter choices, random seeds, feature selection, etc., meaning that comparison on just one setting can unfairly favour certain algorithms. In this work, we show significant variance in fairness achieved by several algorithms and the influence of the learning pipeline on fairness scores. We highlight that most bias mitigation techniques can achieve comparable performance, given the freedom to perform hyperparameter optimization, suggesting that the choice of the evaluation parameters—rather than the mitigation technique itself—can sometimes create the perceived superiority of one method over another. We hope our work encourages future research on how various choices in the lifecycle of developing an algorithm impact fairness, and trends that guide the selection of appropriate algorithms.
\end{abstract}

\section{Introduction}

Over the past decade, concerns about fairness and discrimination in Machine Learning (ML) systems have emerged as critical issues, driving extensive research into the development of fair ML practices, including mitigation algorithms and fairness criteria~\citep{mehrabi2021survey,gohar2023survey, barocas2023fairness}. This has led to emerging global AI regulation focused on mitigating discrimination in AI/ML systems, mandating the reporting of fairness metrics of algorithms in compliance with various anti-discrimination laws such as the disparate impact doctrine~\citep{Civil}. 

\begin{figure*}
    \centering
    \includegraphics[width=0.98\linewidth]{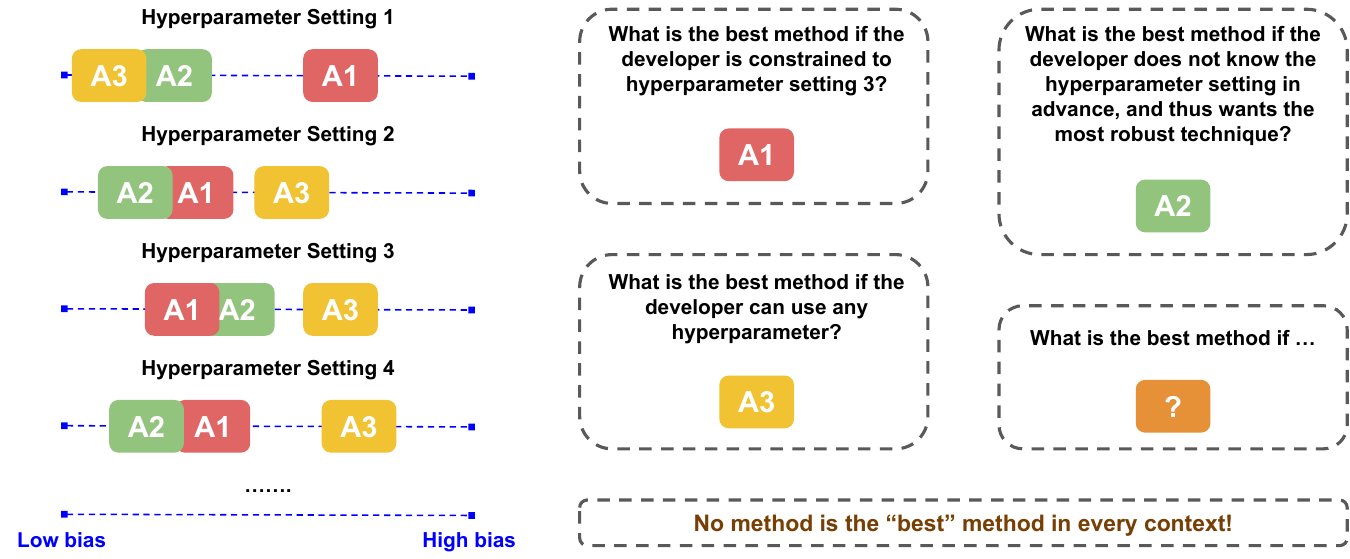}
    \caption{Motivation behind a more nuanced and context-aware benchmarking of bias mitigation techniques, instead of using a uniform evaluation setup or attempting to find the "best" technique.}
    \label{fig:evaluation_problem}
\end{figure*}

However, despite the regulatory efforts, recent research has increasingly shown that the fair ML pipeline suffers from instability and high variance in fairness measures, which can mask the underlying unfairness while creating an illusion of fairness~\citep{black2023toward}. For instance, recent work has pointed out how fairness measures vary across different training runs or between training and deployment, challenging the effectiveness, reliability, and utility of existing methods \citep{baldini2021your,black2021leave,friedler2019comparative,ganesh2023impact}. Additionally, the multitude of mitigation techniques and fairness metrics further complicate accurate benchmarking. Therefore, from both a regulatory perspective and best practices, such variances must be taken into account to accurately represent the performance of these systems and fairness intervention methods.

While recent works have highlighted the issue of variance in fairness, existing fairness benchmarks predominately operate under a single identical training environment (e.g., hyperparameters, random seed, etc.) to ensure more accurate and fair comparisons~\citep{han2023ffb}. However, this fails to consider the sensitivity of fairness to hyperparameter choices, which may mask the nuances of various bias mitigation techniques, favoring one over another~\citep{dooley2024rethinking}, as illustrated in Figure \ref{fig:evaluation_problem}. We postulate that after accounting for the variance in fairness due to the hyperparameter choice, most bias mitigation methods achieve comparable performance. This further raises important questions about the one-dimensional nature of existing fairness evaluations. Is the "best" model simply the one that performs optimally under specific hyperparameter configurations, or should fairness assessments take a more holistic approach? For instance, how do we account for trade-offs between fairness, interpretability, stability, and resource constraints? Should fairness evaluations prioritize consistency over best performance, or should context-specific factors like deployment environments and real-world implications dictate the criteria for success? These questions highlight the need for more nuanced and multidimensional fairness benchmarks beyond traditional measures.

\textbf{Contributions of our work.} We show that bias mitigation algorithms are highly sensitive to several choices made in the learning pipeline. Consequently, without incorporating the broader context, a one-dimensional comparative analysis of these algorithms can create a false sense of fairness. We highlight that most mitigation algorithms can achieve comparable performance under appropriate hyperparameter optimization and, therefore, advocate going beyond the narrow view of the fairness-utility tradeoff to explore other factors that impact model deployment. We conduct a large-scale empirical analysis to support our claims. We hope our work inspires future research that explores the interplay between bias mitigation algorithms and the entire learning pipeline, rather than studying them in isolation, as has often been the case in the literature.


\section{Related Work}

Researchers have developed a range of mitigation techniques and notions to address unfairness in ML systems, targeting different stages of the ML pipeline, including pre-processing, in-processing, and post-processing methods~\citep{mehrabi2021survey,pessach2022review,gohar2024long}. Pre-processing balances data distribution across protected groups, reducing variance, while post-processing modifies model outputs without accessing internal algorithms, ideal for black-box models. In contrast, in-processing methods impose fairness constraints on the model or modify the objective function to mitigate bias. There is inherent randomness in the training process, which, while vital for convergence and generalization~\citep{noh2017regularizing}, can be a source of high-fairness variance. Moreover, these techniques also have control parameters (for example, regularization weight) that must be optimized for the training data, further introducing variance~\citep{bottou2012stochastic}. In this work, we limit our focus on in-processing techniques due to the high variance exhibited by these methods~\citep{baldini2021your,black2021leave,friedler2019comparative,ganesh2023impact,perrone2021fair}.

There is increasing evidence in the literature of the instability of fairness metrics associated with non-determinism in model training and decisions~\citep{black2022model,baldini2021your,friedler2019comparative}. This includes identical training environments with small changes such as different random seeds~\citep{black2021leave}, sampling~\citep{ganesh2023impact}, hyperparameter choices~\citep{ganesh2024empirical,gohar2023towards}, or even differences in train-test-split~\citep{friedler2019comparative}, which have been shown to lead to meaningful differences in group fairness performance across different runs. The issue arises when this instability is used to report higher fairness performance, which calls into question the effectiveness of fairness assessments~\citep{black2024d}. 
A few recent works have explored this issue from a regulatory fairness assessment standpoint where, inadvertently or deliberately, an actor can misrepresent the fairness performance. For instance, a parallel work by ~\citet{simson2024one} proposes using multiverse analysis to keep track of all possible decision combinations for data processing and its impact on model fairness. We build on a similar argument, focusing instead on decisions made during the algorithm design, highlighting the instability of various fairness intervention methods.

Several previous attempts have been made to benchmark bias mitigation algorithms~\citep{bellamy2019ai,bird2020fairlearn,han2023ffb}, including those conducted by papers proposing new mitigation techniques~\citep{kamishima2012fairness,li2022kernel,madras2018learning,adel2019one,zhang2018mitigating}. As highlighted by \citet{han2023ffb}, every paper that proposes a new bias mitigation algorithm often introduces its own experimental setup. This lack of standardization can make it difficult to compare different algorithms effectively. To overcome this issue, \citet{han2023ffb} created the FFB benchmark, offering a comprehensive evaluation over a wide range of datasets and algorithms to allow fair comparison. While FFB provides an excellent foundation to compare and benchmark bias mitigation algorithms, its attempts to find an algorithm that provides the best fairness-utility tradeoff (which they claim is HSIC) comes with two important caveats, (a) FFB only considers a single hyperparameter setting, which, while useful for standardization, overlooks the sensitivity of fairness scores to hyperparameter choices~\citep{perrone2021fair}, and (b) FFB aggregates results across multiple datasets, which can obscure the nuances of performance on individual datasets. In our work, we demonstrate that the fairness variability due to hyperparameter choices can often mask unfairness and raise concerns about existing evaluation techniques. 

\section{Variance in Bias Mitigation}

In this section, we argue that benchmarking under different settings can reveal trends that are lost when sticking to only a single standardized hyperparameter setting or aggregating results across multiple datasets, as done by \citet{han2023ffb}. Thus, a more nuanced approach to benchmarking bias mitigation techniques is needed to capture the strengths and limitations of various algorithms.

\subsection{Experiment setup}

For our experiments, we borrow from \citet{han2023ffb}, using their open-source code~\footnote{\url{https://github.com/ahxt/fair_fairness_benchmark}}. We focus on the seven tabular datasets and the seven bias mitigation algorithms used in their benchmark (not including the standard empirical risk minimization without fairness constraints). In addition to varying the random seed for training and the control parameter for bias mitigation as done in the benchmark, we also vary the batch size, learning rate, and model architecture to explore several different hyperparameter settings. More details on the experiment setup are delegated to Appendix \ref{sec:app_experiment_setup}.

\subsection{Case Study: Adult Dataset}

We start with a case study on the Adult dataset~\citep{adult_2} and show the changing trends across different hyperparameters. We plot fairness as demographic parity and utility as accuracy, studying the fairness-utility tradeoff across various settings in Figure \ref{fig:adult_variance}. Additional discussions on other datasets and fairness metrics are present in Appendix \ref{sec:app_hyperparameter}.

\begin{figure*}
    \centering
    \includegraphics[width=0.98\linewidth]{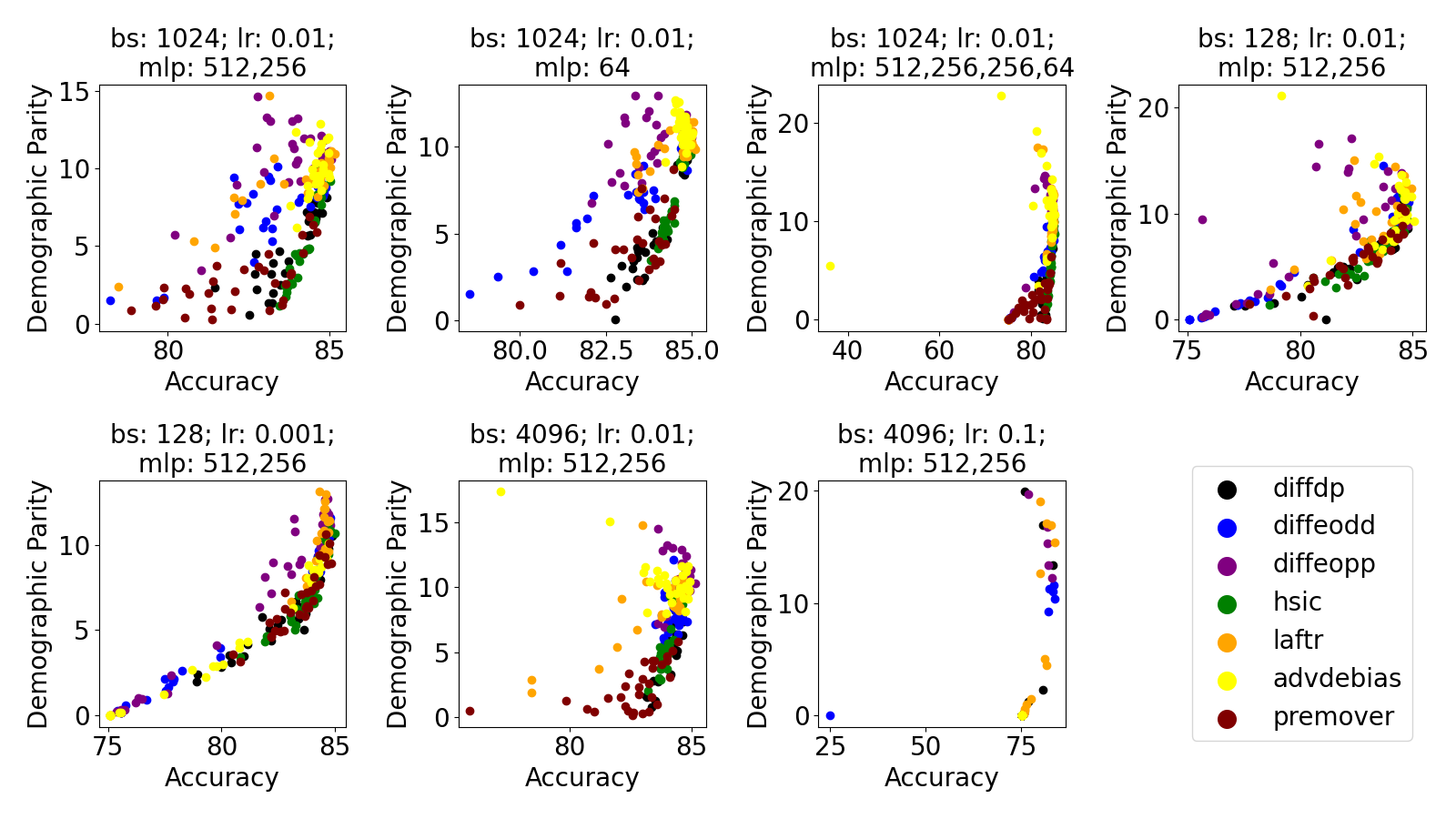}
    \caption{Fairness-utility (demographic parity-accuracy) tradeoff across various settings for the Adult dataset. Each graph represents a different combination of hyperparameters, and each dot in the graph represents a separate training run. Multiple dots for the same mitigation algorithm in the same graph represent runs with changing random seeds and control parameters.}
    \label{fig:adult_variance}
\end{figure*}

We first observe the absence of a clear winner among bias mitigation algorithms; no single method consistently and significantly outperforms the others. While HSIC achieves better tradeoffs under the hyperparameter setting used by \citet{han2023ffb}, other techniques such as PRemover and DiffDP perform equally well -- or sometimes even better -- under other hyperparameter settings. Thus, a comparative analysis limited to just one combination of hyperparameters fails to capture the competitive performance of other algorithms.

Another interesting set of results comes from the hyperparameter setting with a large batch size and a high learning rate. These settings are particularly relevant in scenarios where the emphasis is on rapid convergence and minimizing the number of training steps, even at the cost of performance, for instance, during rapid prototyping, edge computing with limited compute cycles or federated learning. Most bias mitigation methods that performed well in other settings fail to even converge under these conditions. Instead, methods like AdvDebias and LAFTR, which did not stand out in other settings, provide good fairness scores when trained under these constraints.

Our findings highlight the importance of considering a diverse range of choices in the learning pipeline when evaluating bias mitigation techniques, as different algorithms excel under different settings. Further interesting trends can be extracted from the comparative analysis of various algorithms across different hyperparameter settings, left for future work. We now turn to a similarly nuanced comparative analysis of bias mitigation techniques, focusing on changing trends across datasets.

\subsection{Changing Trends Across Datasets}

In the previous section, we observed different techniques perform better than others under changing settings within a single dataset. We now extend this observation to multiple datasets to show that even the trends across different datasets can vary significantly, and the choice of combining results from all datasets, as done by \citet{han2023ffb}, can obscure these trends. We record the fairness (demographic parity) and utility (accuracy) across different datasets and bias mitigation algorithms in Figure \ref{fig:all_variance}. Results for other fairness metrics are present in Appendix \ref{sec:app_datasets}.

\begin{figure*}
    \centering
    \includegraphics[width=0.98\linewidth]{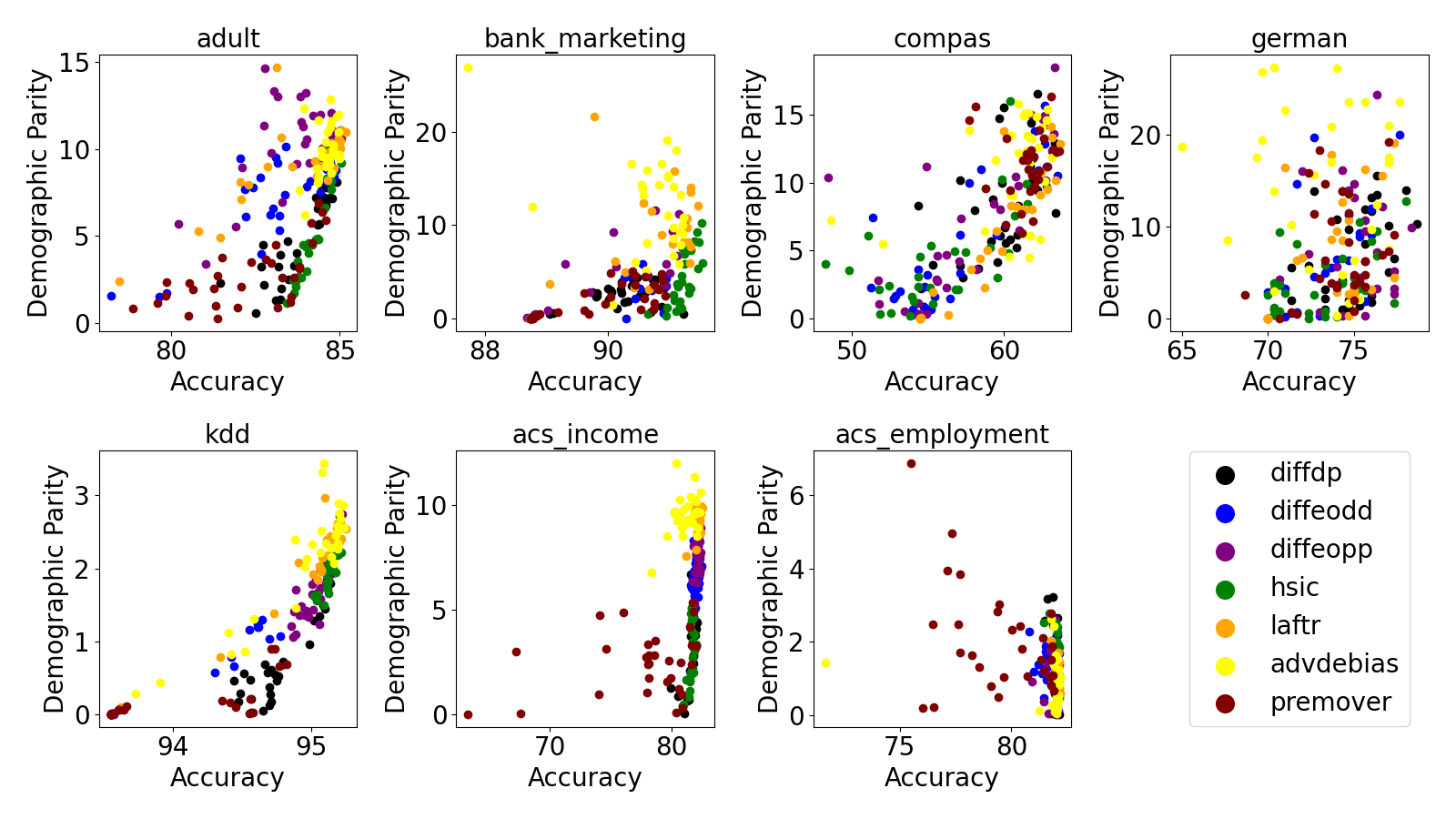}
    \caption{Fairness-utility (demographic parity-accuracy) tradeoff across various datasets, under their default hyperparameters. Each dot in the graph represents a separate training run with changing random seeds and control parameters.}
    \label{fig:all_variance}
\end{figure*}

We begin again with HSIC, which \citet{han2023ffb} identifies as the algorithm offering the best tradeoff overall. While HSIC does well on most datasets, it is noticeably not the unanimous top choice for both the COMPAS and German datasets. Interestingly, both of these are small datasets with smaller batch sizes used for training. Given that HSIC relies on pairwise similarity in a batch to estimate dependence, its superiority with larger batch sizes but its lackluster performance with smaller batch sizes is not surprising. Most of the datasets analyzed by \citet{han2023ffb} were big datasets with large batch sizes. Thus, combining results from multiple datasets overshadowed the trends present only in smaller datasets, effectively hiding HSIC’s shortcomings.

Another intriguing trend is present in the performance of LAFTR, the adversarial representation learning-based bias mitigation technique. Generally, LAFTR underperforms across various datasets, offering poor tradeoffs. However, it performs surprisingly well on the COMPAS dataset, providing competitive tradeoffs against other techniques. This anomaly may be linked to the data preprocessing step adopted by \citet{han2023ffb}, where all categorical features are converted into one-hot encodings. Since LAFTR relies on representation learning at its core, this explosion in the number of features can make the representation learning task more difficult, thus hurting the eventual mitigation attempts. Notably, the COMPAS dataset, where LAFTR excels, contains the least number of categorical features and, consequently, the smallest input size compared to other datasets. Thus, LAFTR’s underperformance might not be simply due to the algorithm itself but rather the choice of input feature representation used.

Finally, we also observe fairness metric-specific trends that affect the comparative analysis between different algorithms. For many datasets like Adult, COMPAS, and KDD, there is a clear tradeoff between demographic parity and accuracy, which isn’t surprising since demographic parity is not aligned with accuracy. In these datasets, we find different mitigation techniques occupy distinct positions in the tradeoff curves. Thus, the choice of the mitigation technique, therefore, depends on the stakeholders’ level of willingness to trade utility for fairness. Conversely, many other datasets do not have these apparent tradeoffs. Here, we find that the results are mixed, with highly overlapping trends across different mitigation algorithms.






\section{Comparisons Beyond the Fairness-Utility Tradeoff}

In the previous section, we discussed the limitations of a generic comparative analysis of bias mitigation algorithms that don't take into account the nuances of the entire learning process. Building on this, we now turn to a practical concern: \textit{choosing the appropriate algorithm}.

We begin this section by first showing that given the opportunity to perform hyperparameter optimization, various mitigation algorithms can provide competitive models. We then discuss how, given the lack of appropriate differentiation between these algorithms in their fairness-utility tradeoff, the selection of the appropriate algorithm can prioritize other factors, like algorithm runtime, complexity, potential robustness, theoretical guarantees, etc. Consequently, selecting the appropriate algorithm and the resulting model would involve balancing these additional considerations, rather than solely focusing on just the best fairness-utility tradeoff.

\subsection{Most Mitigation Algorithms are Competitive}

As we saw in Figure \ref{fig:adult_variance} (and Appendix \ref{sec:app_hyperparameter}), different algorithms do well under varying settings. In several real-world applications, many of these choices are flexible, and hyperparameter optimization plays an important role in model selection. Thus, when comparing different algorithms, it is important to focus on evaluating the best-performing models from each algorithm, as these are the models that would be deployed if those algorithms were used.

To perform this comparison, we only filter the models at the Pareto front for various algorithms after searching through different hyperparameters and random seeds collected in Figure \ref{fig:pareto_front}. Trends for other fairness metrics are present in Appendix \ref{sec:app_competitive}. We find that several algorithms can provide competitive tradeoffs for almost every dataset. For instance, DiffDP, PRemover, and HSIC demonstrate excellent fairness-utility tradeoffs for the Adult dataset, while all seven bias mitigation algorithms exhibit competitive tradeoffs on the German dataset. With multiple algorithms showing similar tradeoffs, it becomes evident that simply evaluating fairness-utility tradeoffs is insufficient when choosing the most suitable bias mitigation technique. We explore these considerations further in the next section.

\begin{figure*}
    \centering
    
    \includegraphics[width=0.98\linewidth]{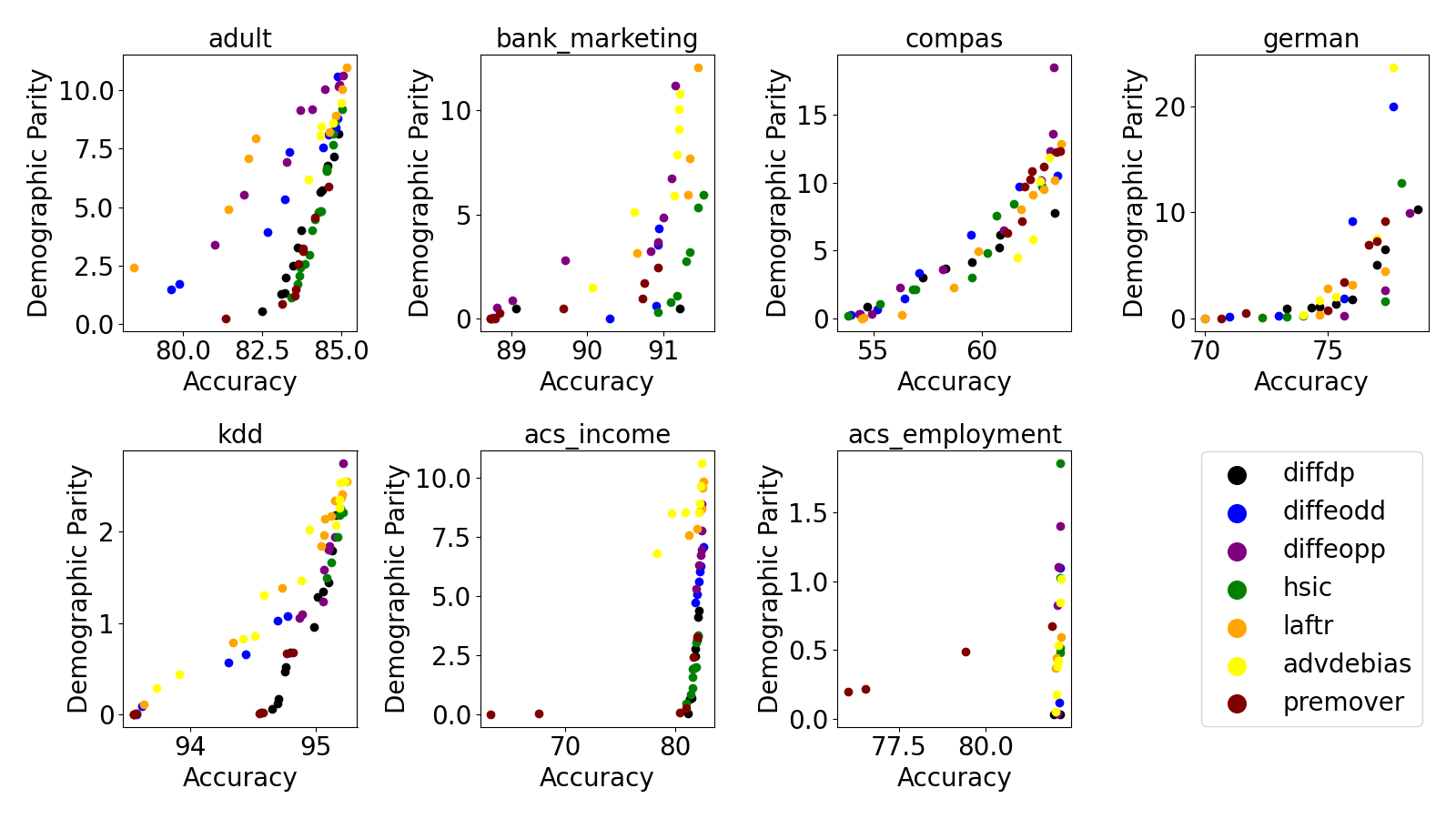}
    \caption{Pareto front of the fairness-utility (demographic parity-accuracy) tradeoff across various datasets. Each dot in the graph represents a separate training run on the pareto front with changing hyperparameters, random seeds and control parameters.}
    \label{fig:pareto_front}
\end{figure*}

\subsection{Choosing the Right Mitigation Technique}

When several bias mitigation algorithms provide similar tradeoffs, selecting one can be challenging. In such cases, additional factors must be considered, such as the specific requirements of the task, the deployment environment, the stakeholders' expectations, etc. Here, we provide some examples of comparisons beyond the fairness-utility tradeoff that can help choose an appropriate algorithm.

\paragraph{Runtime:} An algorithm's runtime can be a crucial factor when comparing bias mitigation techniques. Even minor differences in runtime might become relevant when multiple runs of the same algorithm are needed, for instance, to perform hyperparameter optimization. Our results, detailed in Table \ref{tab:runtime}, reveal interesting trends in training runtime across various algorithms. We find the algorithms HSIC, LAFTR, PRemover, DiffEOdd, and DiffEOpp to be quite expensive, while in contrast, algorithms DiffDP and AdvDebias offer runtime comparable to the standard empirical risk minimization. Considering the competitive tradeoffs achieved by DiffDP, in addition to the lower runtime, it emerges as an appropriate choice for settings where computational efficiency is critical, surpassing other well-performing but slower methods like HSIC and PRemover.

\begin{table}[h!]
    \centering
    \footnotesize
    \renewcommand{\tabcolsep}{3.8pt}
    \begin{tabular}{lrrrrrrrr}
        \toprule
        \multirow{2}{*}{\textbf{Dataset}} & \multicolumn{8}{c}{\textbf{Runtime (rounded to 5s intervals)}} \\
        \cmidrule{2-9}
        & \textbf{ERM} & \textbf{DiffDP} & \textbf{DiffEOdd} & \textbf{DiffEOpp} & \textbf{HSIC} & \textbf{LAFTR} & \textbf{PRemover} & \textbf{AdvDebias} \\
        \midrule
        Bank & 1m 15s & 1m 15s & 1m 45s & 1m 45s & 1m 50s & 1m 46s & 1m 50s & 1m 25s \\
        German & 30s & 30s & 35s & 35s & 40s & 35s & 40s & 30s \\
        Adult & 1m 40s & 1m 40s & 1m 45s & 1m 45s & 2m 0s & 1m 50s & 2m 0s & 1m 40s \\
        COMPAS & 30s & 30s & 30s & 30s & 30s & 30s & 30s & 30s \\
        KDDCensus & 6m 45s & 6m 50s & 10m 40s & 10m 40s & 10m 5s & 10m 0s & 9m 50s & 6m 50s \\
        ACS-I & 7m 10s & 7m 10s & 9m 50s & 9m 50s & 9m 50s & 9m 30s & 10m 0s & 7m 50s \\
        ACS-E & 13m 40s & 13m 45s & 15m 40s & 15m 40s & 16m 20s & 15m 50s & 16m 10s & 13m 40s \\
        ACS-P & 7m 20s & 7m 35s & 9m 50s & 9m 50s & 9m 40s & 10m 5s & 10m 5s & 7m 40s \\
        ACS-M & 4m 40s & 4m 45s & 6m 5s & 6m 5s & 6m 0s & 6m 10s & 6m 0s & 4m 50s \\
        ACS-T & 7m 30s & 7m 30s & 10m 5s & 10m 5s & 10m 15s & 10m 0s & 10m 20s & 8m 0s \\
        \bottomrule
    \end{tabular}
    \vspace{4pt}
    \caption{Training runtime of mitigation algorithms across datasets, under default hyperparameters.}
    \label{tab:runtime}
\end{table}

\paragraph{Theoretical Guarantees and Procedural Requirements:} Another important consideration when selecting the appropriate algorithm is the theoretical guarantees that some techniques can offer. For instance, while adding regularizers to the training objective can be useful, it does not provide any form of guarantee for the model's final fairness scores. In contrast, methods like HSIC and LAFTR can provide theoretical bounds on the fairness of the final model, albeit limited to only simpler models~\citep{li2022kernel,madras2018learning}. 

Furthermore, the deployed models may need to comply with specific procedural requirements, which can influence the choice of the mitigation algorithm. For instance, one might need to choose between algorithms focusing on outcome fairness (such as DiffDP, DiffEOpp, DiffEOdd) versus those focusing on process fairness (such as HSIC, LAFTR, PRemover, AdvDebias). The specific requirements of the application can dictate the choice of the algorithm, looking beyond the tradeoffs it can provide.

\paragraph{Multiplicity and Arbitrariness:} Model multiplicity refers to the existence of a set of good models, which have similar performance but differ in their predictions for individuals~\citep{marx2020predictive,black2022model}. Existing works have shown that bias mitigation can exacerbate multiplicity concerns, leading to arbitrariness in individual-level predictions~\citep{long2024individual}. However, the degree of multiplicity introduced can vary depending on the mitigation algorithm used. Following \citet{long2024individual}, we define the set of competing models as models with similar accuracy under ERM and record multiplicity using ambiguity~\citep{marx2020predictive}, which is the fraction of data points whose predictions change across different models within the set of good models, in Table \ref{tab:ambiguity}.

\begin{table}[h!]
    \centering
    \footnotesize
    \renewcommand{\tabcolsep}{4pt}
    \begin{tabular}{lrrrrrrrr}
        \toprule
        \multirow{2}{*}{\textbf{Dataset}} & \multicolumn{8}{c}{\textbf{Ambiguity}} \\
        \cmidrule{2-9}
        & \textbf{ERM} & \textbf{DiffDP} & \textbf{DiffEOdd} & \textbf{DiffEOpp} & \textbf{HSIC} & \textbf{LAFTR} & \textbf{PRemover} & \textbf{AdvDebias} \\
        \midrule
        Bank & 0.15 & 0.16 & 0.16 & 0.18 & 0.17 & 0.19 & 0.15 & 0.26 \\
        German & 0.55 & 0.57 & 0.57 & 0.63 & 0.55 & 0.59 & 0.60 & 0.87 \\
        Adult & 0.17 & 0.30 & 0.37 & 0.42 & 0.28 & 0.32 & 0.34 & 0.47 \\
        COMPAS & 0.93 & 0.99 & 1.0 & 1.0 & 1.0 & 1.0 & 0.92 & 0.99 \\
        KDDCensus & 0.04 & 0.06 & 0.04 & 0.06 & 0.05 & 0.06 & 0.04 & 0.09 \\
        ACS-I & 0.26 & 0.35 & 0.38 & 0.35 & 0.38 & 0.32 & 0.75 & 0.49 \\
        ACS-E & 0.14 & 0.20 & 0.30 & 0.21 & 0.26 & 0.20 & 0.49 & 0.37 \\
        ACS-P & 0.27 & 0.33 & 0.38 & 0.39 & 0.34 & 0.45 & 0.32 & 0.69 \\
        ACS-M & 0.26 & 0.29 & 0.27 & 0.29 & 0.31 & 0.38 & 0.20 & 0.61 \\
        ACS-T & 0.70 & 0.81 & 0.88 & 0.80 & 0.79 & 0.91 & 0.92 & 0.90 \\
        \bottomrule
    \end{tabular}
    \vspace{4pt}
    \caption{Ambiguity scores of mitigation algorithms across datasets, under default hyperparameters.}
    \label{tab:ambiguity}
\end{table}

Unsurprisingly, most bias mitigation techniques exhibit higher ambiguity than ERM, which aligns with the observations made by \citet{long2024individual}. However, an interesting exception is the PRemover algorithm, which achieves remarkably low ambiguity scores across many datasets, distinguishing it from other algorithms. Strikingly, at the same time, PRemover also shows significantly high ambiguity in several other datasets, highlighting its behavior on both extremes. Thus, for certain datasets, PRemover could be considered a superior choice compared to other methods like HSIC and DiffDP, which, while offering similar trade-offs, tend to introduce more arbitrariness into the model. In contrast to PRemover, the AdvDebias algorithm consistently results in very high ambiguity scores, making it a poor choice in contexts where minimizing arbitrariness is crucial.

In this section, we showed several examples of additional factors to consider when selecting an algorithm for a specific use case. Naturally, this list is not exhaustive, as additional considerations may arise depending on the specific application context. The objective of our study was to emphasize the lack of distinction between mitigation algorithms that focus solely on the fairness-utility tradeoff and the importance of choosing algorithms that offer additional advantages beyond this tradeoff. With these results, we hope to move away from the narrative of a single optimal bias mitigation technique and emphasize the need for context-dependent comparative analysis.

\section{Discussion}

In this paper, we underscore the limitations of current fairness benchmarking practices that rely on uniform evaluation setups. We demonstrate that hyperparameter optimization can yield similar performance across different bias mitigation techniques, raising questions about the effectiveness of existing benchmarks and the criteria for selecting appropriate fairness algorithms.

\textbf{Context-dependent evaluation.} We argue that the current one-dimensional approach to fairness evaluation may be insufficient. Given the high variability in fairness scores, relying on a single run or, conversely, simply aggregating multiple training runs, both common practices across different dimensions, may not always provide an appropriate comparison of bias mitigation techniques. 

For example, when models are too large and retraining is impractical, choosing fairness interventions that prioritize stability and consistent scores may be more appropriate. On the other hand, if sufficient computational resources exist to explore hyperparameter options, selecting the best-performing model might be more valid. Additionally, explainability, runtime, and scalability constraints can significantly impact the choice of fairness assessments. Ultimately, the method of comparing algorithms depends on the context. However, in all cases, it is crucial to consider the variability introduced by hyperparameter tuning.

\textbf{Future work.} Our experiments were limited to in-processing techniques in bias mitigation. In the future, we plan to explore a broader range of methods, including pre and post-processing. Moreover, we have not explored the potential presence of consistent fairness trends for different hyperparameter choices covered in the experiments. It would be interesting to investigate whether we can identify patterns that guide our decisions to choose better hyperparameter settings for various bias mitigation algorithms. Finally, while evidence in the literature would suggest similar trends exist even with hyperparameters in other parts of the pipeline, for instance, data processing~\citep{simson2024one}, our empirical results are limited to hyperparameter choices during training. Further work on a large-scale study of the impact of various choices in the lifetime of an algorithm design is needed.

\ack

Funding support for project activities has been partially provided by the Canada CIFAR AI Chair, FRQNT scholarship, and NSERC discovery award. We also express our gratitude to Compute Canada and Mila clusters for their support in providing facilities for our evaluations.
Lu Cheng is supported by the National Science Foundation (NSF) Grant \#2312862, NIH \#R01AG091762, and a Cisco gift grant.

\bibliographystyle{abbrvnat}
\bibliography{references}

\begin{thebibliography}{41}
\providecommand{\natexlab}[1]{#1}
\providecommand{\url}[1]{\texttt{#1}}
\expandafter\ifx\csname urlstyle\endcsname\relax
  \providecommand{\doi}[1]{doi: #1}\else
  \providecommand{\doi}{doi: \begingroup \urlstyle{rm}\Url}\fi

\bibitem[cen(2000)]{census-income_(kdd)_117}
{Census-Income (KDD)}.
\newblock UCI Machine Learning Repository, 2000.
\newblock {DOI}: https://doi.org/10.24432/C5N30T.

\bibitem[Adel et~al.(2019)Adel, Valera, Ghahramani, and Weller]{adel2019one}
T.~Adel, I.~Valera, Z.~Ghahramani, and A.~Weller.
\newblock One-network adversarial fairness.
\newblock In \emph{Proceedings of the AAAI Conference on Artificial Intelligence}, volume~33, pages 2412--2420, 2019.

\bibitem[Baharlouei et~al.()Baharlouei, Nouiehed, Beirami, and Razaviyayn]{baharloueirenyi}
S.~Baharlouei, M.~Nouiehed, A.~Beirami, and M.~Razaviyayn.
\newblock R{\'e}nyi fair inference.
\newblock In \emph{International Conference on Learning Representations}.

\bibitem[Baldini et~al.(2021)Baldini, Wei, Ramamurthy, Yurochkin, and Singh]{baldini2021your}
I.~Baldini, D.~Wei, K.~N. Ramamurthy, M.~Yurochkin, and M.~Singh.
\newblock Your fairness may vary: Pretrained language model fairness in toxic text classification.
\newblock \emph{arXiv preprint arXiv:2108.01250}, 2021.

\bibitem[Barocas et~al.(2023)Barocas, Hardt, and Narayanan]{barocas2023fairness}
S.~Barocas, M.~Hardt, and A.~Narayanan.
\newblock \emph{Fairness and machine learning: Limitations and opportunities}.
\newblock MIT Press, 2023.

\bibitem[Becker and Kohavi(1996)]{adult_2}
B.~Becker and R.~Kohavi.
\newblock {Adult}.
\newblock UCI Machine Learning Repository, 1996.
\newblock {DOI}: https://doi.org/10.24432/C5XW20.

\bibitem[Bellamy et~al.(2019)Bellamy, Dey, Hind, Hoffman, Houde, Kannan, Lohia, Martino, Mehta, Mojsilovi{\'c}, et~al.]{bellamy2019ai}
R.~K. Bellamy, K.~Dey, M.~Hind, S.~C. Hoffman, S.~Houde, K.~Kannan, P.~Lohia, J.~Martino, S.~Mehta, A.~Mojsilovi{\'c}, et~al.
\newblock Ai fairness 360: An extensible toolkit for detecting and mitigating algorithmic bias.
\newblock \emph{IBM Journal of Research and Development}, 63\penalty0 (4/5):\penalty0 4--1, 2019.

\bibitem[Beutel et~al.(2017)Beutel, Chen, Zhao, and Chi]{beutel2017data}
A.~Beutel, J.~Chen, Z.~Zhao, and E.~H. Chi.
\newblock Data decisions and theoretical implications when adversarially learning fair representations.
\newblock \emph{arXiv preprint arXiv:1707.00075}, 2017.

\bibitem[Bird et~al.(2020)Bird, Dud{\'\i}k, Edgar, Horn, Lutz, Milan, Sameki, Wallach, and Walker]{bird2020fairlearn}
S.~Bird, M.~Dud{\'\i}k, R.~Edgar, B.~Horn, R.~Lutz, V.~Milan, M.~Sameki, H.~Wallach, and K.~Walker.
\newblock Fairlearn: A toolkit for assessing and improving fairness in ai.
\newblock \emph{Microsoft, Tech. Rep. MSR-TR-2020-32}, 2020.

\bibitem[Black and Fredrikson(2021)]{black2021leave}
E.~Black and M.~Fredrikson.
\newblock Leave-one-out unfairness.
\newblock In \emph{Proceedings of the 2021 ACM Conference on Fairness, Accountability, and Transparency}, pages 285--295, 2021.

\bibitem[Black et~al.(2022)Black, Raghavan, and Barocas]{black2022model}
E.~Black, M.~Raghavan, and S.~Barocas.
\newblock Model multiplicity: Opportunities, concerns, and solutions.
\newblock In \emph{Proceedings of the 2022 ACM Conference on Fairness, Accountability, and Transparency}, pages 850--863, 2022.

\bibitem[Black et~al.(2023)Black, Naidu, Ghani, Rodolfa, Ho, and Heidari]{black2023toward}
E.~Black, R.~Naidu, R.~Ghani, K.~Rodolfa, D.~Ho, and H.~Heidari.
\newblock Toward operationalizing pipeline-aware ml fairness: A research agenda for developing practical guidelines and tools.
\newblock In \emph{Proceedings of the 3rd ACM Conference on Equity and Access in Algorithms, Mechanisms, and Optimization}, pages 1--11, 2023.

\bibitem[Black et~al.(2024)Black, Gillis, and Hall]{black2024d}
E.~Black, T.~Gillis, and Z.~Y. Hall.
\newblock D-hacking.
\newblock In \emph{The 2024 ACM Conference on Fairness, Accountability, and Transparency}, pages 602--615, 2024.

\bibitem[Bottou(2012)]{bottou2012stochastic}
L.~Bottou.
\newblock Stochastic gradient descent tricks.
\newblock In \emph{Neural Networks: Tricks of the Trade: Second Edition}, pages 421--436. Springer, 2012.

\bibitem[Ding et~al.(2021)Ding, Hardt, Miller, and Schmidt]{ding2021retiring}
F.~Ding, M.~Hardt, J.~Miller, and L.~Schmidt.
\newblock Retiring adult: New datasets for fair machine learning.
\newblock \emph{Advances in neural information processing systems}, 34:\penalty0 6478--6490, 2021.

\bibitem[Dooley et~al.(2024)Dooley, Sukthanker, Dickerson, White, Hutter, and Goldblum]{dooley2024rethinking}
S.~Dooley, R.~Sukthanker, J.~Dickerson, C.~White, F.~Hutter, and M.~Goldblum.
\newblock Rethinking bias mitigation: Fairer architectures make for fairer face recognition.
\newblock \emph{Advances in Neural Information Processing Systems}, 36, 2024.

\bibitem[Edwards and Storkey(2016)]{edwards2016censoring}
H.~Edwards and A.~Storkey.
\newblock Censoring representations with an adversary.
\newblock In \emph{4th International Conference on Learning Representations}, pages 1--14, 2016.

\bibitem[Friedler et~al.(2019)Friedler, Scheidegger, Venkatasubramanian, Choudhary, Hamilton, and Roth]{friedler2019comparative}
S.~A. Friedler, C.~Scheidegger, S.~Venkatasubramanian, S.~Choudhary, E.~P. Hamilton, and D.~Roth.
\newblock A comparative study of fairness-enhancing interventions in machine learning.
\newblock In \emph{Proceedings of the conference on fairness, accountability, and transparency}, pages 329--338, 2019.

\bibitem[Ganesh(2024)]{ganesh2024empirical}
P.~Ganesh.
\newblock An empirical investigation into benchmarking model multiplicity for trustworthy machine learning: A case study on image classification.
\newblock In \emph{Proceedings of the IEEE/CVF Winter Conference on Applications of Computer Vision}, pages 4488--4497, 2024.

\bibitem[Ganesh et~al.(2023)Ganesh, Chang, Strobel, and Shokri]{ganesh2023impact}
P.~Ganesh, H.~Chang, M.~Strobel, and R.~Shokri.
\newblock On the impact of machine learning randomness on group fairness.
\newblock In \emph{Proceedings of the 2023 ACM Conference on Fairness, Accountability, and Transparency}, pages 1789--1800, 2023.

\bibitem[Gohar and Cheng(2023)]{gohar2023survey}
U.~Gohar and L.~Cheng.
\newblock A survey on intersectional fairness in machine learning: Notions, mitigation, and challenges.
\newblock \emph{arXiv preprint arXiv:2305.06969}, 2023.

\bibitem[Gohar et~al.(2023)Gohar, Biswas, and Rajan]{gohar2023towards}
U.~Gohar, S.~Biswas, and H.~Rajan.
\newblock Towards understanding fairness and its composition in ensemble machine learning.
\newblock In \emph{2023 IEEE/ACM 45th International Conference on Software Engineering (ICSE)}, pages 1533--1545. IEEE, 2023.

\bibitem[Gohar et~al.(2024)Gohar, Tang, Wang, Zhang, Spirtes, Liu, and Cheng]{gohar2024long}
U.~Gohar, Z.~Tang, J.~Wang, K.~Zhang, P.~L. Spirtes, Y.~Liu, and L.~Cheng.
\newblock Long-term fairness inquiries and pursuits in machine learning: A survey of notions, methods, and challenges.
\newblock \emph{arXiv preprint arXiv:2406.06736}, 2024.

\bibitem[Gretton et~al.(2005)Gretton, Bousquet, Smola, and Sch{\"o}lkopf]{gretton2005measuring}
A.~Gretton, O.~Bousquet, A.~Smola, and B.~Sch{\"o}lkopf.
\newblock Measuring statistical dependence with hilbert-schmidt norms.
\newblock In \emph{International conference on algorithmic learning theory}, pages 63--77. Springer, 2005.

\bibitem[Han et~al.(2023)Han, Chi, Chen, Wang, Zhao, Zou, and Hu]{han2023ffb}
X.~Han, J.~Chi, Y.~Chen, Q.~Wang, H.~Zhao, N.~Zou, and X.~Hu.
\newblock Ffb: A fair fairness benchmark for in-processing group fairness methods.
\newblock In \emph{International Conference on Learning Representations}. ICLR, 2023.

\bibitem[Hofmann(1994)]{statlog_(german_credit_data)_144}
H.~Hofmann.
\newblock {Statlog (German Credit Data)}.
\newblock UCI Machine Learning Repository, 1994.
\newblock {DOI}: https://doi.org/10.24432/C5NC77.

\bibitem[Justice.(2023)]{Civil}
U.~S. D.~O. Justice.
\newblock Title vi legal manual, section vii: Proving discrimination – disparate impact., Oct 2023.
\newblock URL \url{https://www.justice.gov/crt/fcs/T6Manual7}.

\bibitem[Kamishima et~al.(2012)Kamishima, Akaho, Asoh, and Sakuma]{kamishima2012fairness}
T.~Kamishima, S.~Akaho, H.~Asoh, and J.~Sakuma.
\newblock Fairness-aware classifier with prejudice remover regularizer.
\newblock In \emph{Machine Learning and Knowledge Discovery in Databases: European Conference, ECML PKDD 2012, Bristol, UK, September 24-28, 2012. Proceedings, Part II 23}, pages 35--50. Springer, 2012.

\bibitem[Larson et~al.(2016)Larson, Mattu, Kirchner, and Angwin]{larson2016propublica}
J.~Larson, S.~Mattu, L.~Kirchner, and J.~Angwin.
\newblock Propublica compas analysis—data and analysis for ‘machine bias.’.
\newblock \emph{https://github.com/propublica/compas-analysis}, 2016.

\bibitem[Li et~al.(2022)Li, P{\'e}rez-Suay, Camps-Valls, and Sejdinovic]{li2022kernel}
Z.~Li, A.~P{\'e}rez-Suay, G.~Camps-Valls, and D.~Sejdinovic.
\newblock Kernel dependence regularizers and gaussian processes with applications to algorithmic fairness.
\newblock \emph{Pattern Recognition}, 132:\penalty0 108922, 2022.

\bibitem[Long et~al.(2024)Long, Hsu, Alghamdi, and Calmon]{long2024individual}
C.~Long, H.~Hsu, W.~Alghamdi, and F.~Calmon.
\newblock Individual arbitrariness and group fairness.
\newblock \emph{Advances in Neural Information Processing Systems}, 36, 2024.

\bibitem[Louppe et~al.(2017)Louppe, Kagan, and Cranmer]{louppe2017learning}
G.~Louppe, M.~Kagan, and K.~Cranmer.
\newblock Learning to pivot with adversarial networks.
\newblock \emph{Advances in neural information processing systems}, 30, 2017.

\bibitem[Madras et~al.(2018)Madras, Creager, Pitassi, and Zemel]{madras2018learning}
D.~Madras, E.~Creager, T.~Pitassi, and R.~Zemel.
\newblock Learning adversarially fair and transferable representations.
\newblock In \emph{International Conference on Machine Learning}, pages 3384--3393. PMLR, 2018.

\bibitem[Marx et~al.(2020)Marx, Calmon, and Ustun]{marx2020predictive}
C.~Marx, F.~Calmon, and B.~Ustun.
\newblock Predictive multiplicity in classification.
\newblock In \emph{International Conference on Machine Learning}, pages 6765--6774. PMLR, 2020.

\bibitem[Mehrabi et~al.(2021)Mehrabi, Morstatter, Saxena, Lerman, and Galstyan]{mehrabi2021survey}
N.~Mehrabi, F.~Morstatter, N.~Saxena, K.~Lerman, and A.~Galstyan.
\newblock A survey on bias and fairness in machine learning.
\newblock \emph{ACM computing surveys (CSUR)}, 54\penalty0 (6):\penalty0 1--35, 2021.

\bibitem[Moro et~al.(2014)Moro, Rita, and Cortez]{bank_marketing_222}
S.~Moro, P.~Rita, and P.~Cortez.
\newblock {Bank Marketing}.
\newblock UCI Machine Learning Repository, 2014.
\newblock {DOI}: https://doi.org/10.24432/C5K306.

\bibitem[Noh et~al.(2017)Noh, You, Mun, and Han]{noh2017regularizing}
H.~Noh, T.~You, J.~Mun, and B.~Han.
\newblock Regularizing deep neural networks by noise: Its interpretation and optimization.
\newblock \emph{Advances in neural information processing systems}, 30, 2017.

\bibitem[Perrone et~al.(2021)Perrone, Donini, Zafar, Schmucker, Kenthapadi, and Archambeau]{perrone2021fair}
V.~Perrone, M.~Donini, M.~B. Zafar, R.~Schmucker, K.~Kenthapadi, and C.~Archambeau.
\newblock Fair bayesian optimization.
\newblock In \emph{Proceedings of the 2021 AAAI/ACM Conference on AI, Ethics, and Society}, pages 854--863, 2021.

\bibitem[Pessach and Shmueli(2022)]{pessach2022review}
D.~Pessach and E.~Shmueli.
\newblock A review on fairness in machine learning.
\newblock \emph{ACM Computing Surveys (CSUR)}, 55\penalty0 (3):\penalty0 1--44, 2022.

\bibitem[Simson et~al.(2024)Simson, Pfisterer, and Kern]{simson2024one}
J.~Simson, F.~Pfisterer, and C.~Kern.
\newblock One model many scores: Using multiverse analysis to prevent fairness hacking and evaluate the influence of model design decisions.
\newblock In \emph{The 2024 ACM Conference on Fairness, Accountability, and Transparency}, pages 1305--1320, 2024.

\bibitem[Zhang et~al.(2018)Zhang, Lemoine, and Mitchell]{zhang2018mitigating}
B.~H. Zhang, B.~Lemoine, and M.~Mitchell.
\newblock Mitigating unwanted biases with adversarial learning.
\newblock In \emph{Proceedings of the 2018 AAAI/ACM Conference on AI, Ethics, and Society}, pages 335--340, 2018.

\end{thebibliography}


\newpage
\appendix

\section{Additional Details on Experiment Setup}
\label{sec:app_experiment_setup}

As we directly borrow the experiment setup from \citet{han2023ffb}, we redirect the reader to their work and the FFB benchmark code~\footnote{\url{https://github.com/ahxt/fair_fairness_benchmark}} for details on the underlying setup. In this section, we briefly mention the datasets and algorithms used in the benchmark, and the new additions and changes we made to their setup.

\subsection{Datasets}

We use 7 different tabular datasets for our experiments. This includes the Adult dataset~\citep{adult_2}, COMPAS dataset ~\citep{larson2016propublica}, German dataset~\citep{statlog_(german_credit_data)_144}, Bank Marketing dataset~\citep{bank_marketing_222}, KDD Census dataset~\citep{census-income_(kdd)_117}, and ACS dataset with tasks Income and Employment~\citep{ding2021retiring}. We use the sensitive attribute \textit{Race} for all datasets, except the Bank Marketing dataset and the German dataset, where we use \textit{Age} as the sensitive attribute.

\subsection{Bias Mitigation Algorithms}

We use 7 different bias mitigation algorithms in our setup. This includes DiffDP, DiffEOdd, DiffEOpp, PRemover~\citep{kamishima2012fairness}, HSIC~\citep{baharloueirenyi,gretton2005measuring,li2022kernel}, AdvDebias~\citep{adel2019one,beutel2017data,edwards2016censoring,louppe2017learning,zhang2018mitigating}, and LAFTR~\citep{madras2018learning}.

\subsection{Hyperparameters}

We use the Adam optimizer, with no weight decay and a step learning rate scheduler for training. We train the model for 150 epochs and record the fairness and accuracy scores at the final epoch.

We use three different values of the control parameter for each algorithm, as defined in Table \ref{tab:control_param}.

\begin{table}[h!]
    \centering
    \footnotesize
    \begin{tabular}{ll}
        \toprule
        \textbf{Algorithm} & \textbf{Control Hyperparameter} \\
        \midrule
        DiffDP & 0.2, 1.0, 1.8 \\
        DiffEOdd & 0.2, 1.0, 1.8 \\
        DiffEOdd & 0.2, 1.0, 1.8 \\
        PRemover & 0.05, 0.25, 0.45 \\
        HSIC & 50, 250, 450 \\
        AdvDebias & 0.2, 1.0, 1.8 \\
        LAFTR & 0.1, 0.5, 4.0 \\
        \bottomrule
    \end{tabular}
    \vspace{4pt}
    \caption{Control hyperparameters.}
    \label{tab:control_param}
\end{table}

We use seven different hyperparameter settings for each dataset, as defined in Table \ref{tab:hyperparameters}.

\begin{table}[h!]
    \centering
    \footnotesize
    \begin{tabular}{lll|lll}
        \toprule
        \multicolumn{3}{c}{Adult and Bank Marketing} & \multicolumn{3}{c}{COMPAS and German} \\
        \midrule
        \textbf{Batch Size} & \textbf{Learning Rate} & \textbf{MLP Layers} & \textbf{Batch Size} & \textbf{Learning Rate} & \textbf{MLP Layers} \\
        \midrule
        1024 & 0.01 & 512,256 & 32 & 0.01 & 512,256 \\
        1024 & 0.01 & 64 & 32 & 0.01 & 64 \\
        1024 & 0.01 & 512,256,256,64 & 32 & 0.01 & 512,256,256,64 \\
        128 & 0.01 & 512,256 & 8 & 0.01 & 512,256 \\
        128 & 0.001 & 512,256 & 8 & 0.001 & 512,256 \\
        4096 & 0.01 & 512,256 & 128 & 0.01 & 512,256 \\
        4096 & 0.1 & 512,256 & 128 & 0.1 & 512,256 \\
        \midrule
        \multicolumn{3}{c}{KDD and ACS} \\
        \midrule
        \textbf{Batch Size} & \textbf{Learning Rate} & \textbf{MLP Layers} \\
        \midrule
        4096 & 0.01 & 512,256 \\
        4096 & 0.01 & 64 \\
        4096 & 0.01 & 512,256,256,64 \\
        512 & 0.01 & 512,256 & \\
        512 & 0.001 & 512,256 & \\
        8192 & 0.01 & 512,256 \\
        8192 & 0.1 & 512,256 \\
        \bottomrule
    \end{tabular}
    \vspace{4pt}
    \caption{Hyperparameters.}
    \label{tab:hyperparameters}
\end{table}

\section{Additional Results for Trends Under Changing Hyperparameters}
\label{sec:app_hyperparameter}

\begin{figure*}
    \centering
    \includegraphics[width=0.98\linewidth]{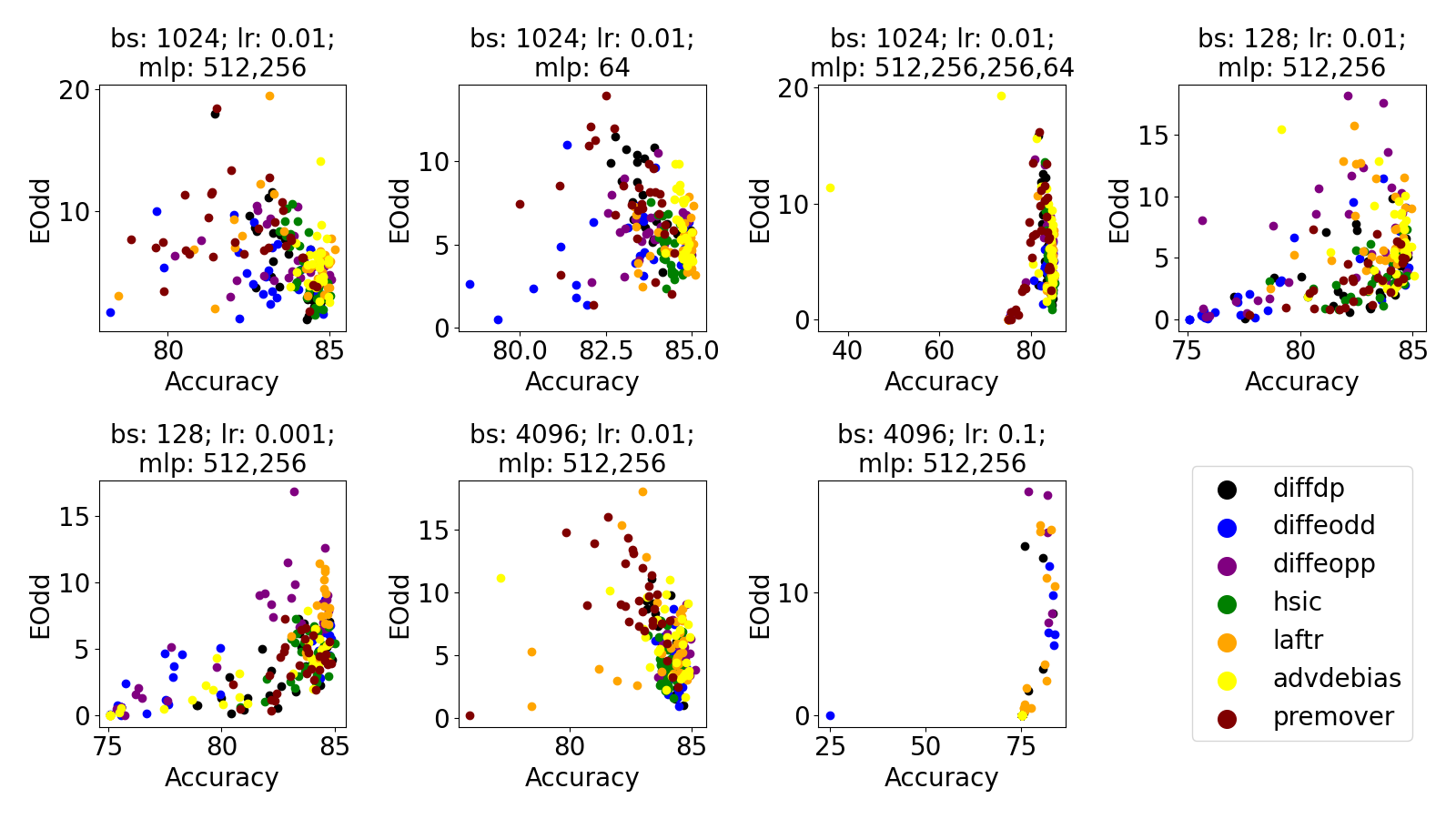}
    \caption{Fairness-utility (equalized odds-accuracy) tradeoff across various settings for the Adult dataset. Each graph represents a different combination of hyperparameters, and each dot in the graph represents a separate training run. Multiple dots for the same mitigation algorithm in the same graph represent runs with changing random seeds and control parameters.}
    \label{fig:adult_variance_eodd}
\end{figure*}

\begin{figure*}
    \centering
    \includegraphics[width=0.98\linewidth]{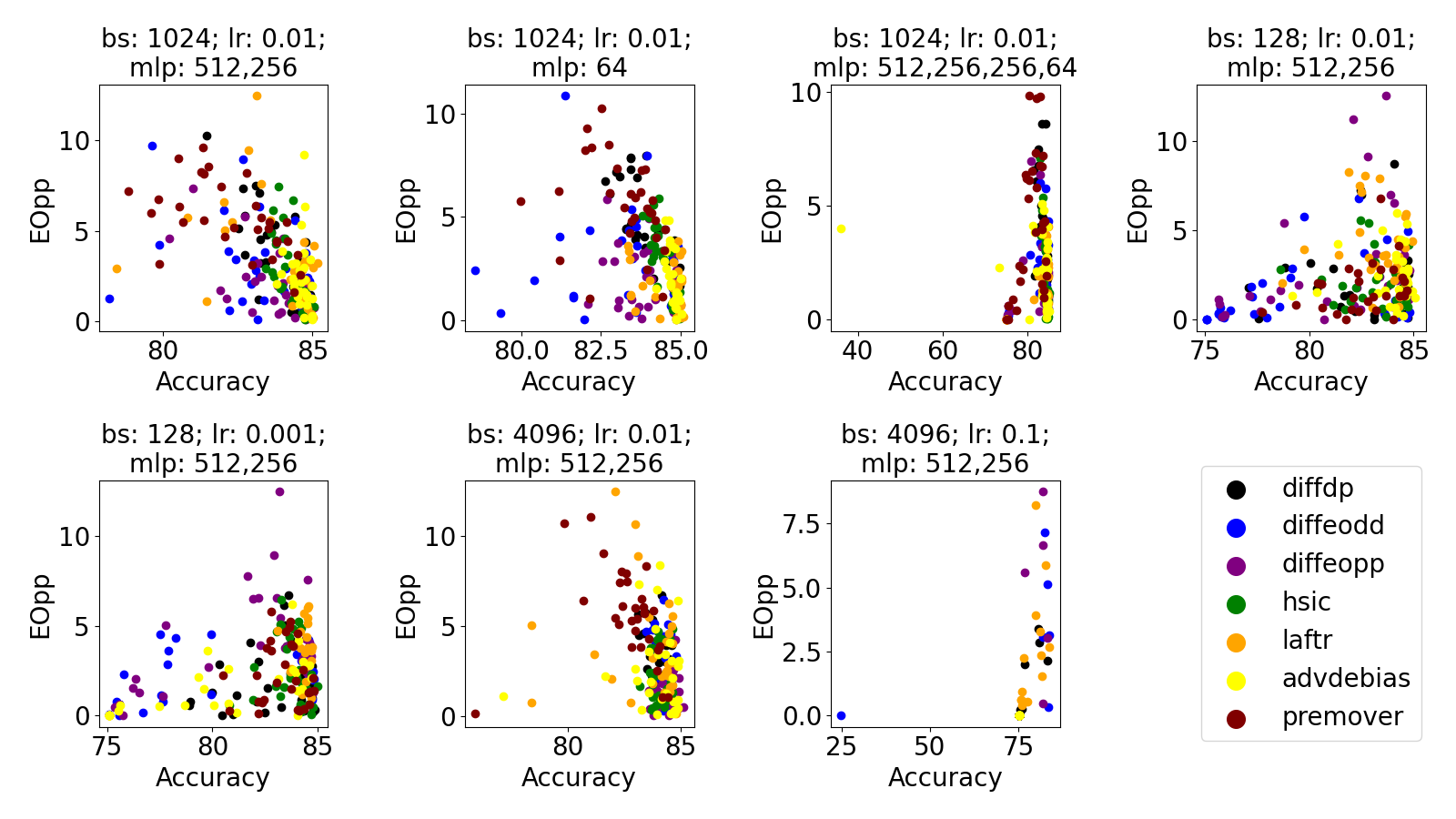}
    \caption{Fairness-utility (equal opportunity-accuracy) tradeoff across various settings for the Adult dataset. Each graph represents a different combination of hyperparameters, and each dot in the graph represents a separate training run. Multiple dots for the same mitigation algorithm in the same graph represent runs with changing random seeds and control parameters.}
    \label{fig:adult_variance_eopp}
\end{figure*}

\begin{figure*}
    \centering
    \includegraphics[width=0.98\linewidth]{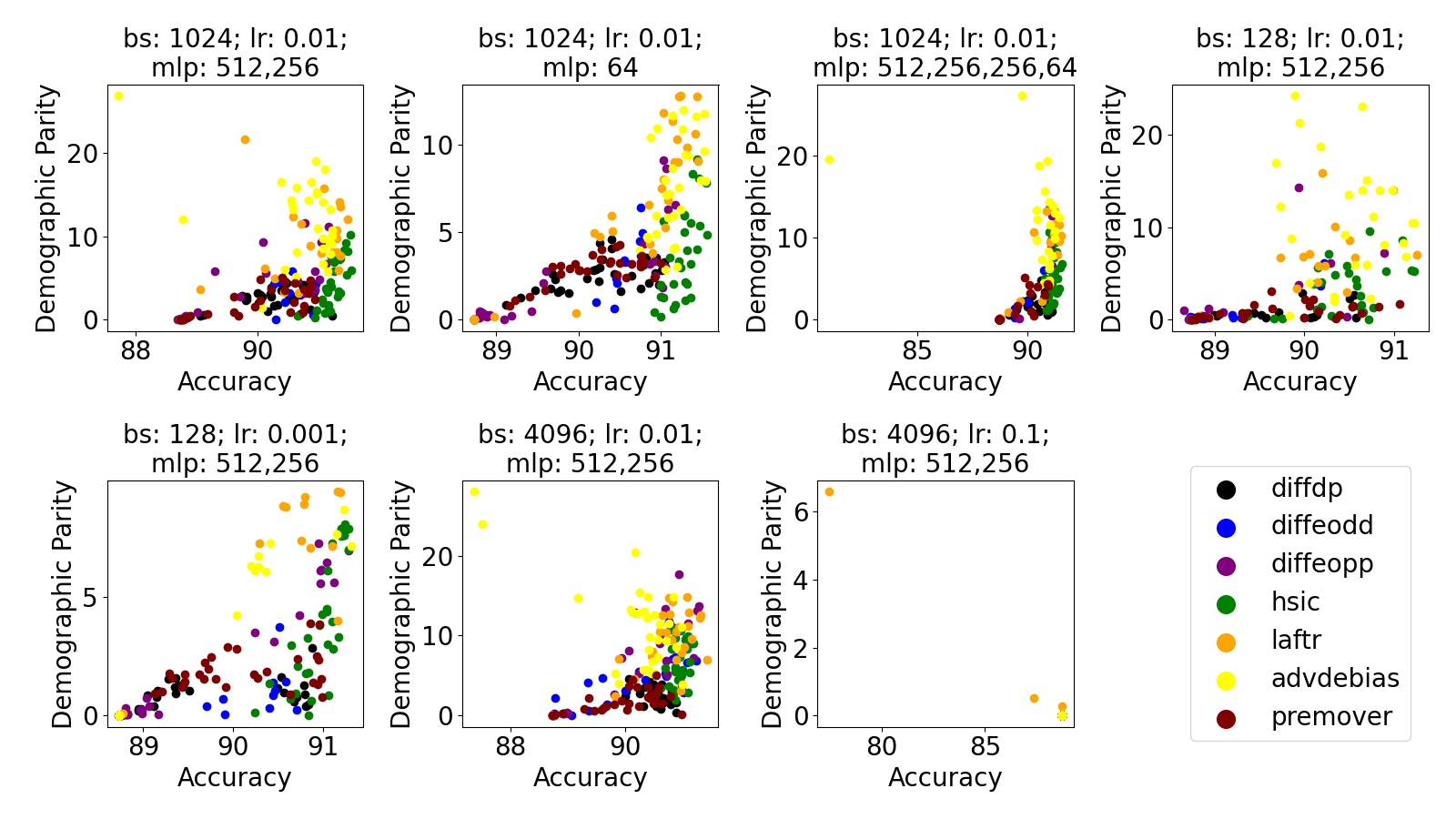}
    \caption{Fairness-utility (demographic parity-accuracy) tradeoff across various settings for the Bank Marketing dataset. Each graph represents a different combination of hyperparameters, and each dot in the graph represents a separate training run. Multiple dots for the same mitigation algorithm in the same graph represent runs with changing random seeds and control parameters.}
    \label{fig:bank_marketing_variance_dp}
\end{figure*}

\begin{figure*}
    \centering
    \includegraphics[width=0.98\linewidth]{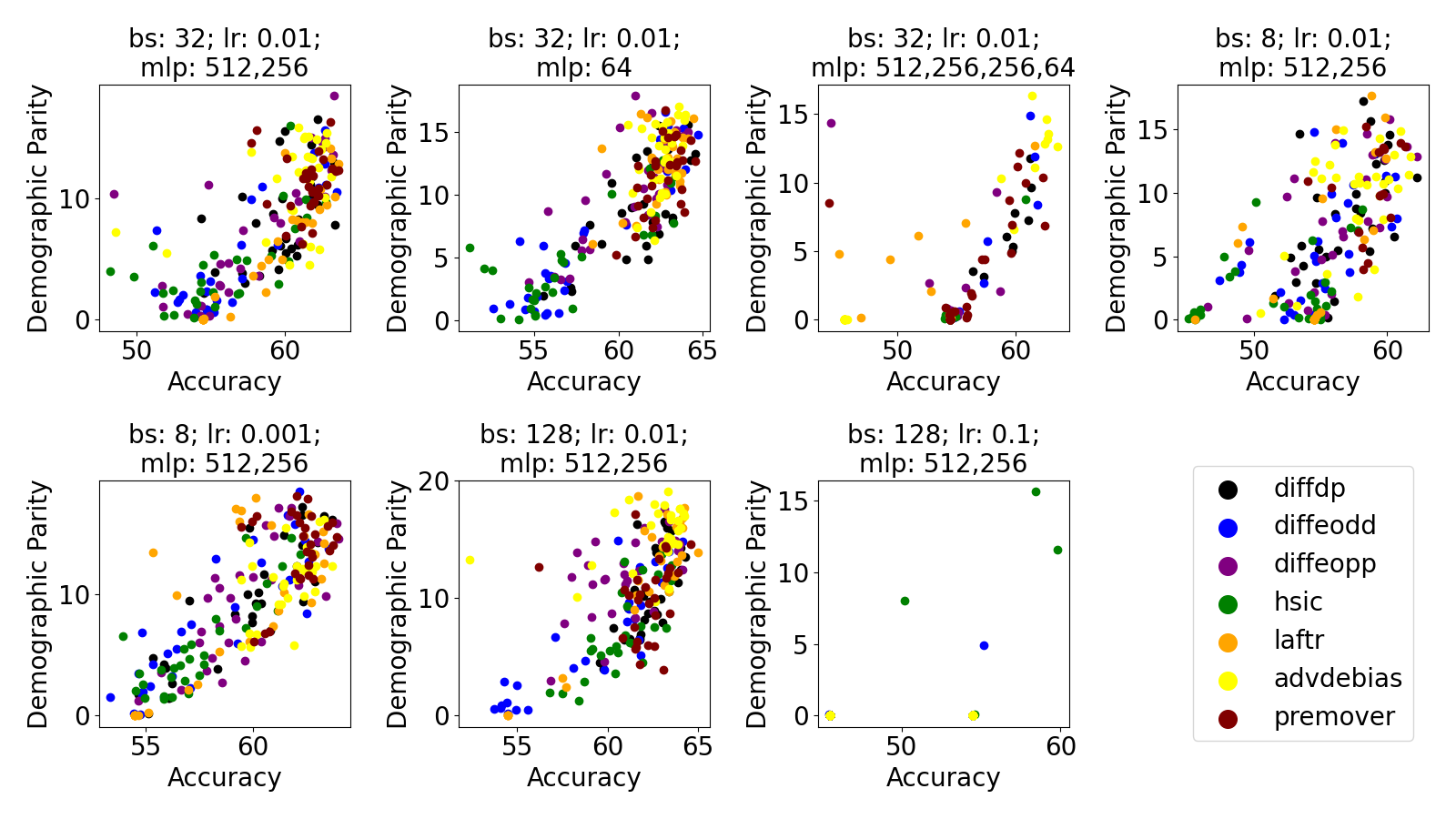}
    \caption{Fairness-utility (demographic parity-accuracy) tradeoff across various settings for the COMPAS dataset. Each graph represents a different combination of hyperparameters, and each dot in the graph represents a separate training run. Multiple dots for the same mitigation algorithm in the same graph represent runs with changing random seeds and control parameters.}
    \label{fig:compas_variance_dp}
\end{figure*}

\begin{figure*}
    \centering
    \includegraphics[width=0.98\linewidth]{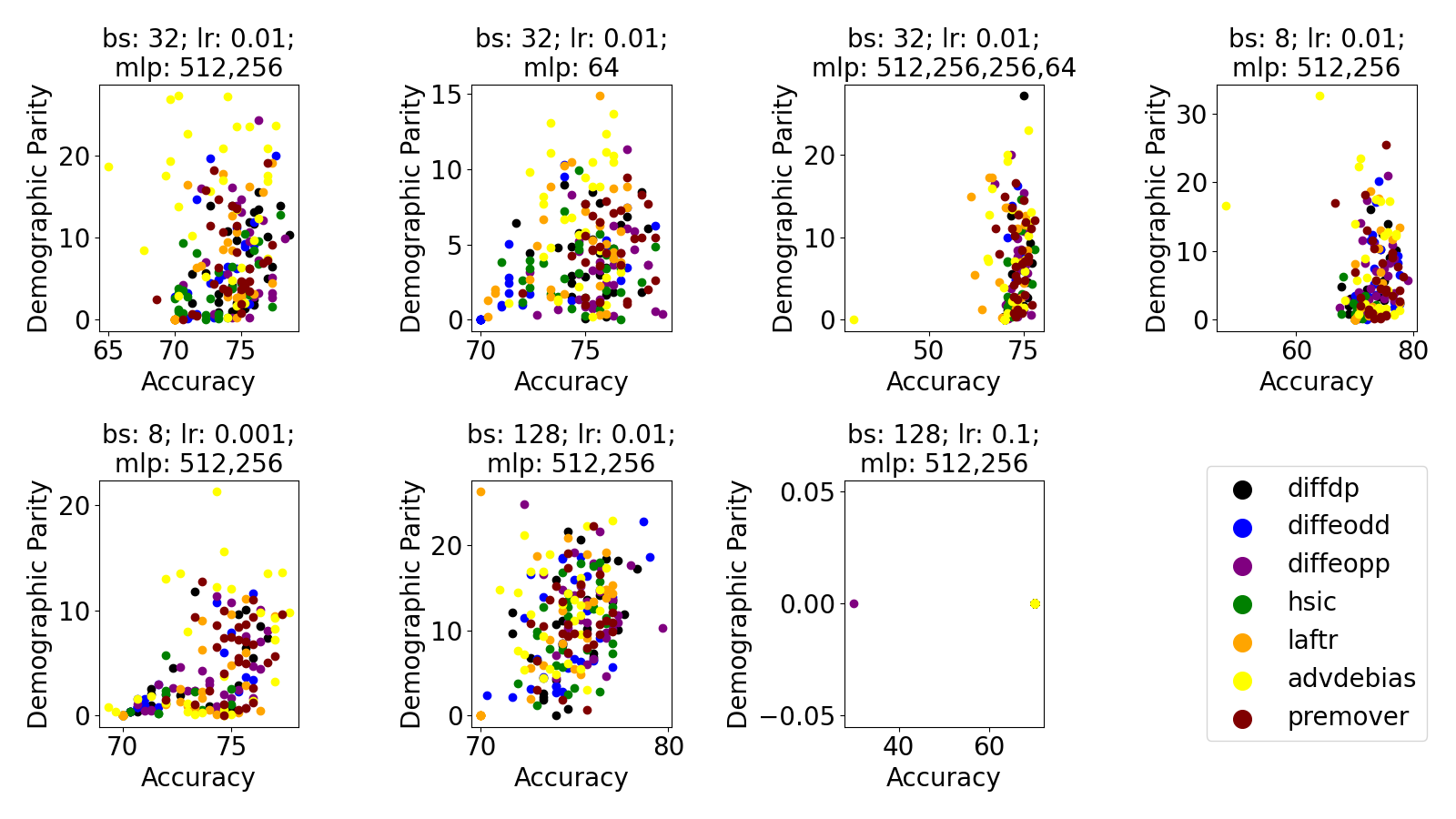}
    \caption{Fairness-utility (demographic parity-accuracy) tradeoff across various settings for the German dataset. Each graph represents a different combination of hyperparameters, and each dot in the graph represents a separate training run. Multiple dots for the same mitigation algorithm in the same graph represent runs with changing random seeds and control parameters.}
    \label{fig:german_variance_dp}
\end{figure*}

\begin{figure*}
    \centering
    \includegraphics[width=0.98\linewidth]{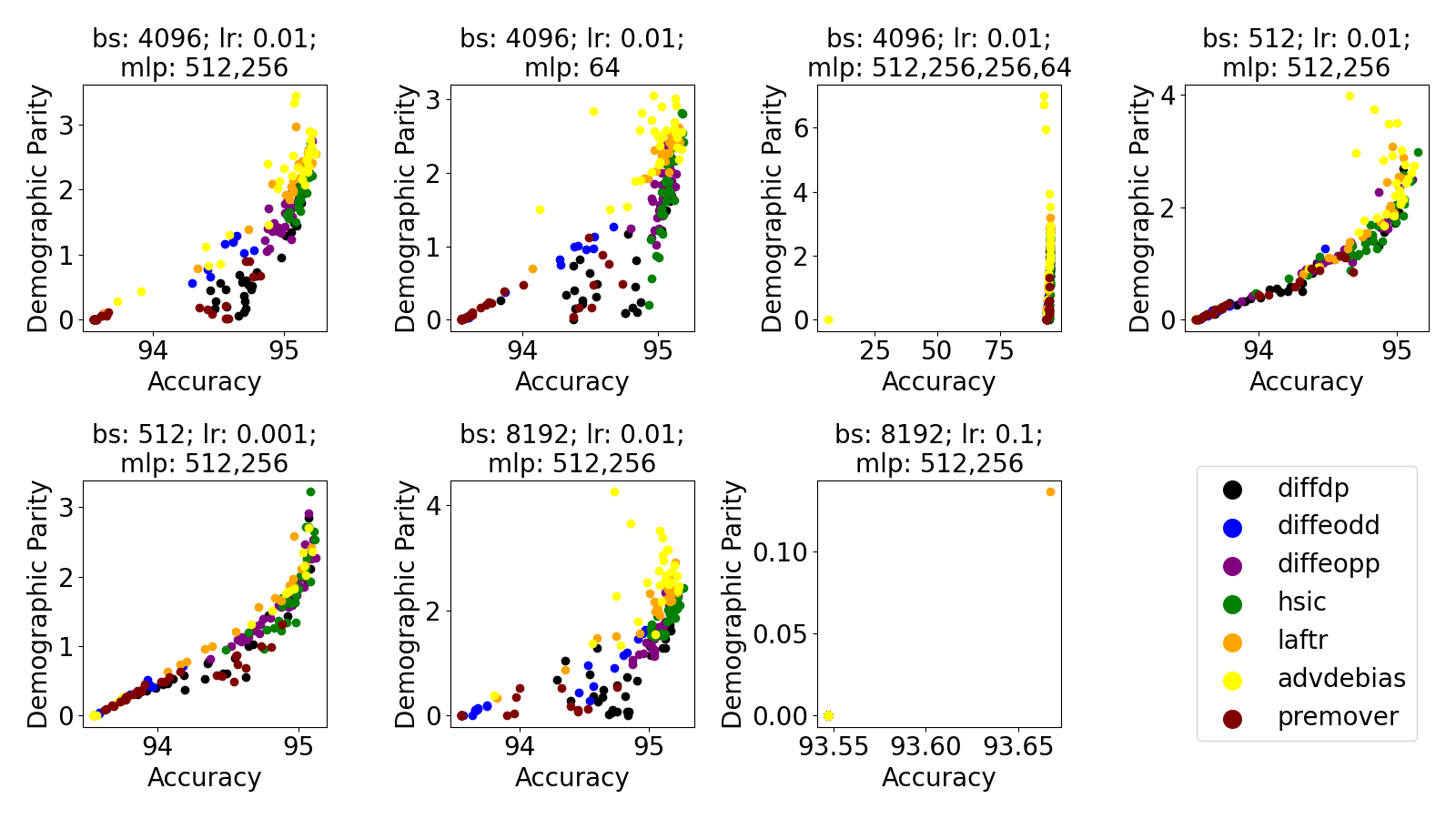}
    \caption{Fairness-utility (demographic parity-accuracy) tradeoff across various settings for the KDDCensus dataset. Each graph represents a different combination of hyperparameters, and each dot in the graph represents a separate training run. Multiple dots for the same mitigation algorithm in the same graph represent runs with changing random seeds and control parameters.}
    \label{fig:kdd_variance_dp}
\end{figure*}

\begin{figure*}
    \centering
    \includegraphics[width=0.98\linewidth]{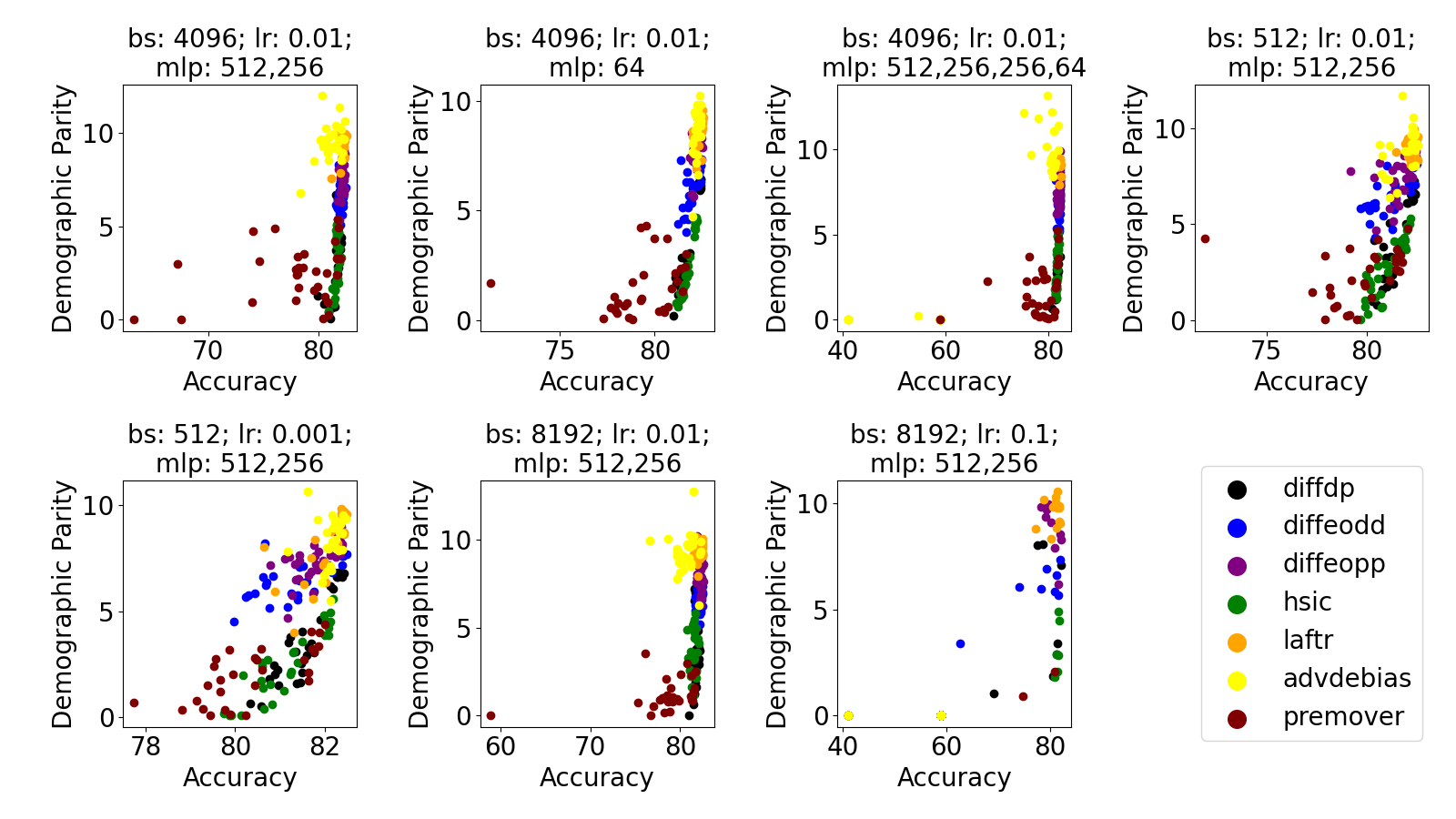}
    \caption{Fairness-utility (demographic parity-accuracy) tradeoff across various settings for the ACS-Income dataset. Each graph represents a different combination of hyperparameters, and each dot in the graph represents a separate training run. Multiple dots for the same mitigation algorithm in the same graph represent runs with changing random seeds and control parameters.}
    \label{fig:income_variance_dp}
\end{figure*}

\begin{figure*}
    \centering
    \includegraphics[width=0.98\linewidth]{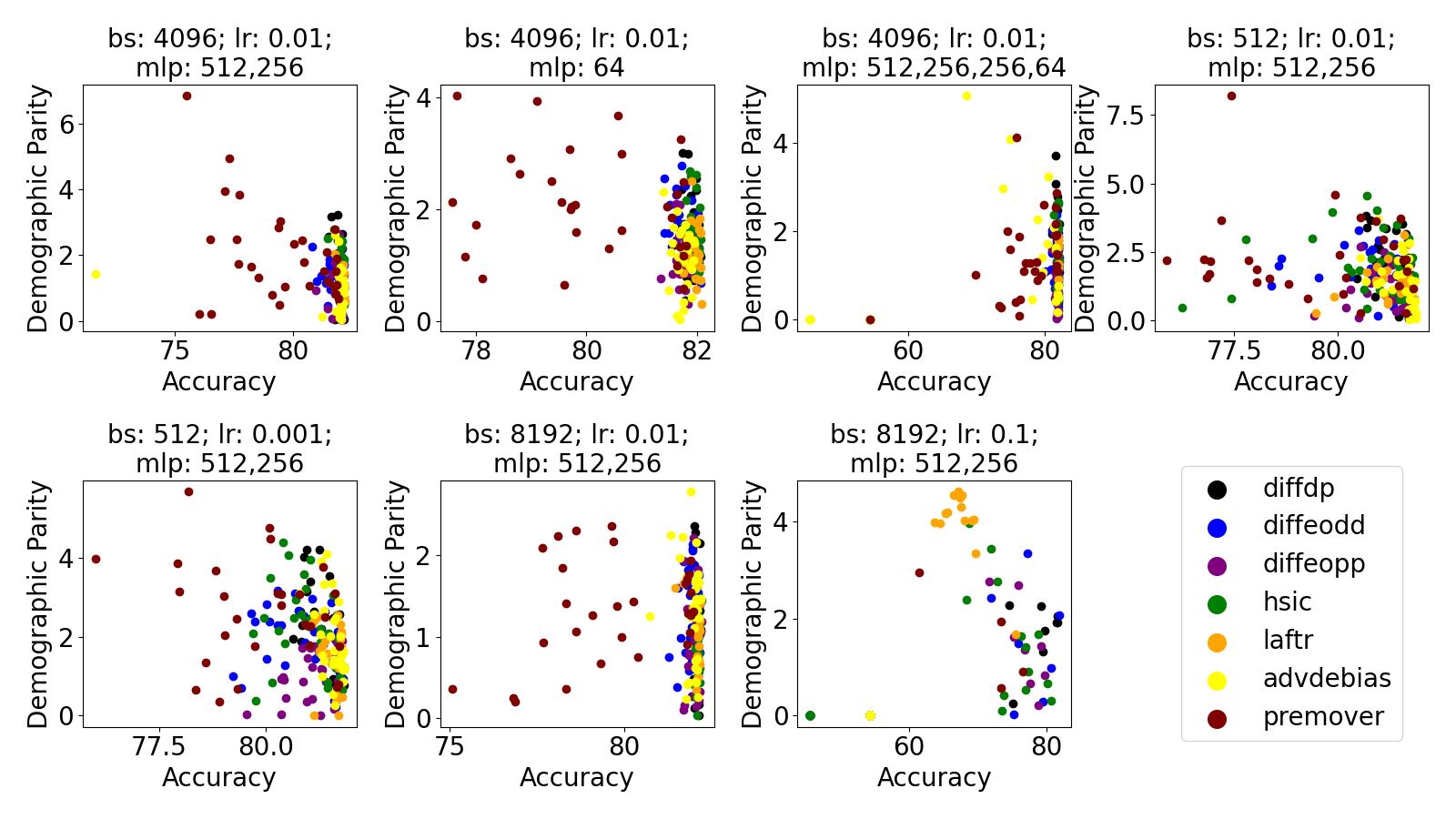}
    \caption{Fairness-utility (demographic parity-accuracy) tradeoff across various settings for the ACS-Employment dataset. Each graph represents a different combination of hyperparameters, and each dot in the graph represents a separate training run. Multiple dots for the same mitigation algorithm in the same graph represent runs with changing random seeds and control parameters.}
    \label{fig:employment_variance_dp}
\end{figure*}

We present additional results for comparing trends under different hyperparameters in the Adult dataset for fairness definitions of equalized odds (Figure \ref{fig:adult_variance_eodd}) and equal opportunity (Figure \ref{fig:adult_variance_eopp}). We also present additional results for comparing trends in other datasets like Bank Marketing dataset (Figure \ref{fig:bank_marketing_variance_dp}), COMPAS dataset (Figure \ref{fig:compas_variance_dp}), German dataset (Figure \ref{fig:german_variance_dp}), KDDCensus dataset (Figure \ref{fig:kdd_variance_dp}), ACS-Income dataset (Figure \ref{fig:income_variance_dp}) and ACS-Employment dataset (Figure \ref{fig:employment_variance_dp}).

\section{Additional Results for Changing Trends Across Datasets}
\label{sec:app_datasets}

\begin{figure*}
    \centering
    \includegraphics[width=0.98\linewidth]{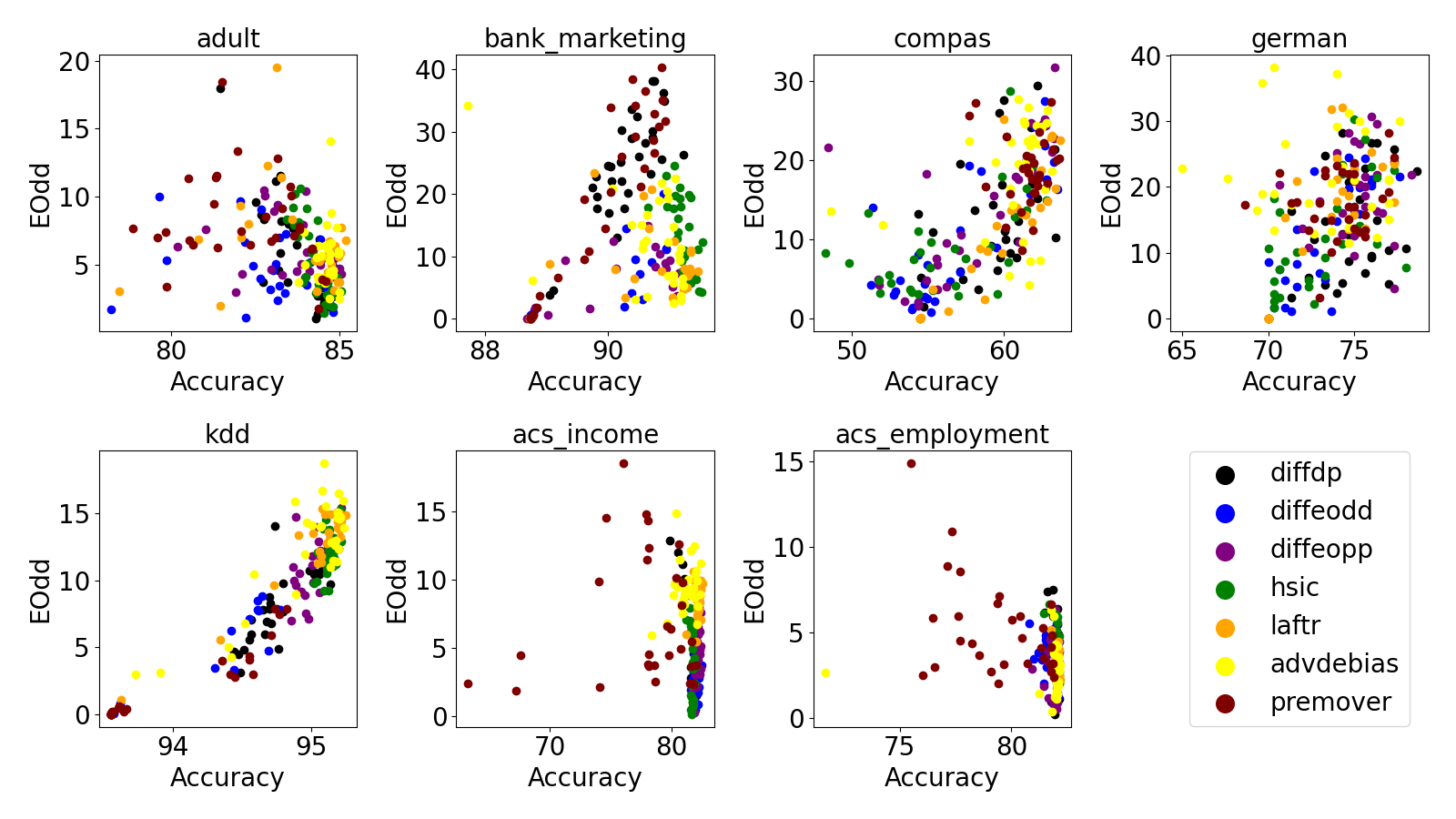}
    \caption{Fairness-utility (equalized odds-accuracy) tradeoff across various datasets, under their default hyperparameters. Each dot in the graph represents a separate training run with changing random seeds and control parameters.}
    \label{fig:all_variance_eodd}
\end{figure*}

\begin{figure*}
    \centering
    \includegraphics[width=0.98\linewidth]{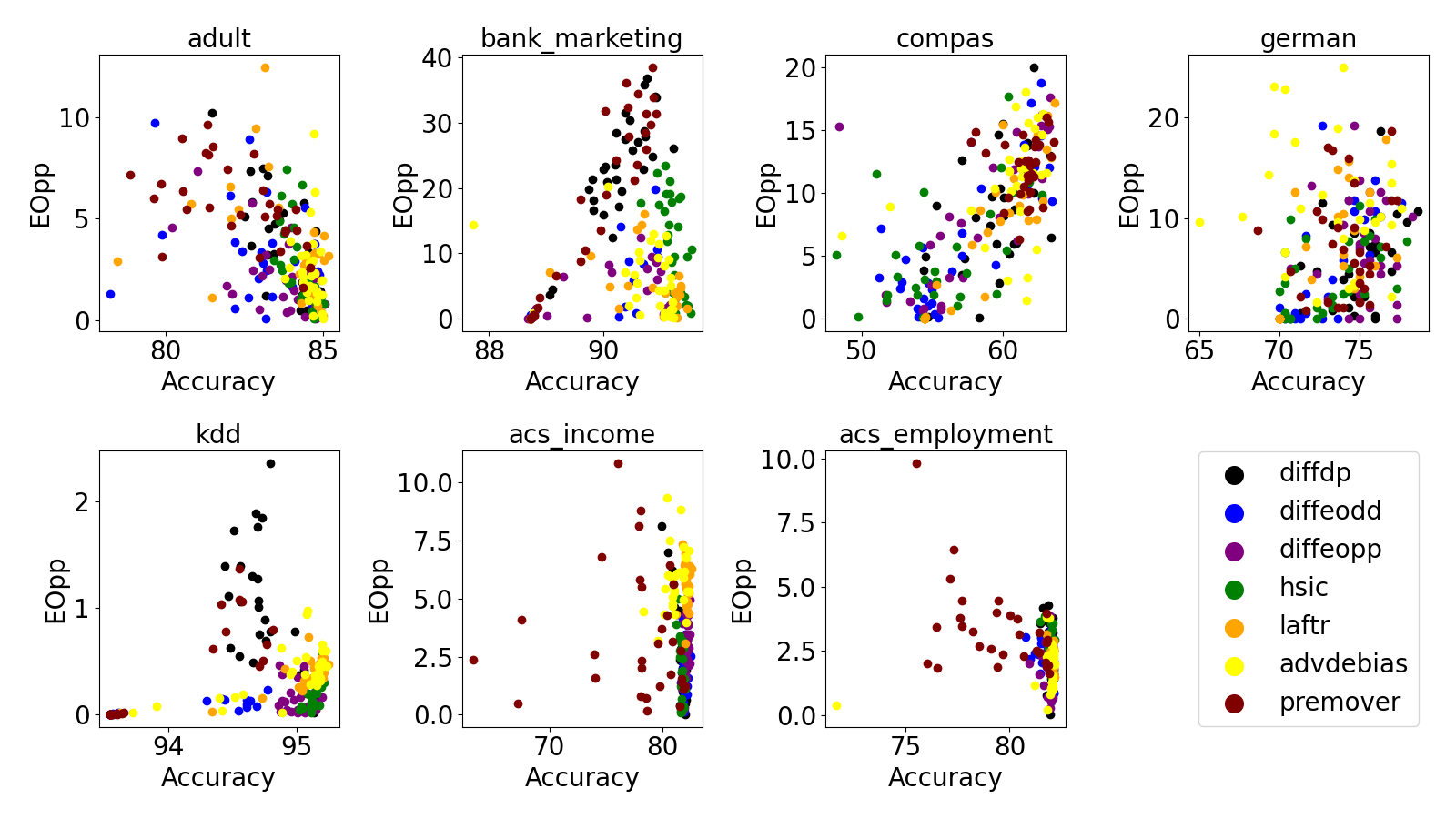}
    \caption{Fairness-utility (equal opportunity-accuracy) tradeoff across various datasets, under their default hyperparameters. Each dot in the graph represents a separate training run with changing random seeds and control parameters.}
    \label{fig:all_variance_eopp}
\end{figure*}

We present additional results for comparing trends across multiple datasets, under fairness definition as equalized odds (Figure \ref{fig:all_variance_eodd}) and equal opportunity (Figure \ref{fig:all_variance_eopp}). Similar to the observations in the main paper, we find distinct trends across different datasets and no clear single bias mitigation algorithm that excels across all datasets.

\section{Additional Results at Pareto Front}
\label{sec:app_competitive}

\begin{figure*}
    \centering
    
    \includegraphics[width=0.98\linewidth]{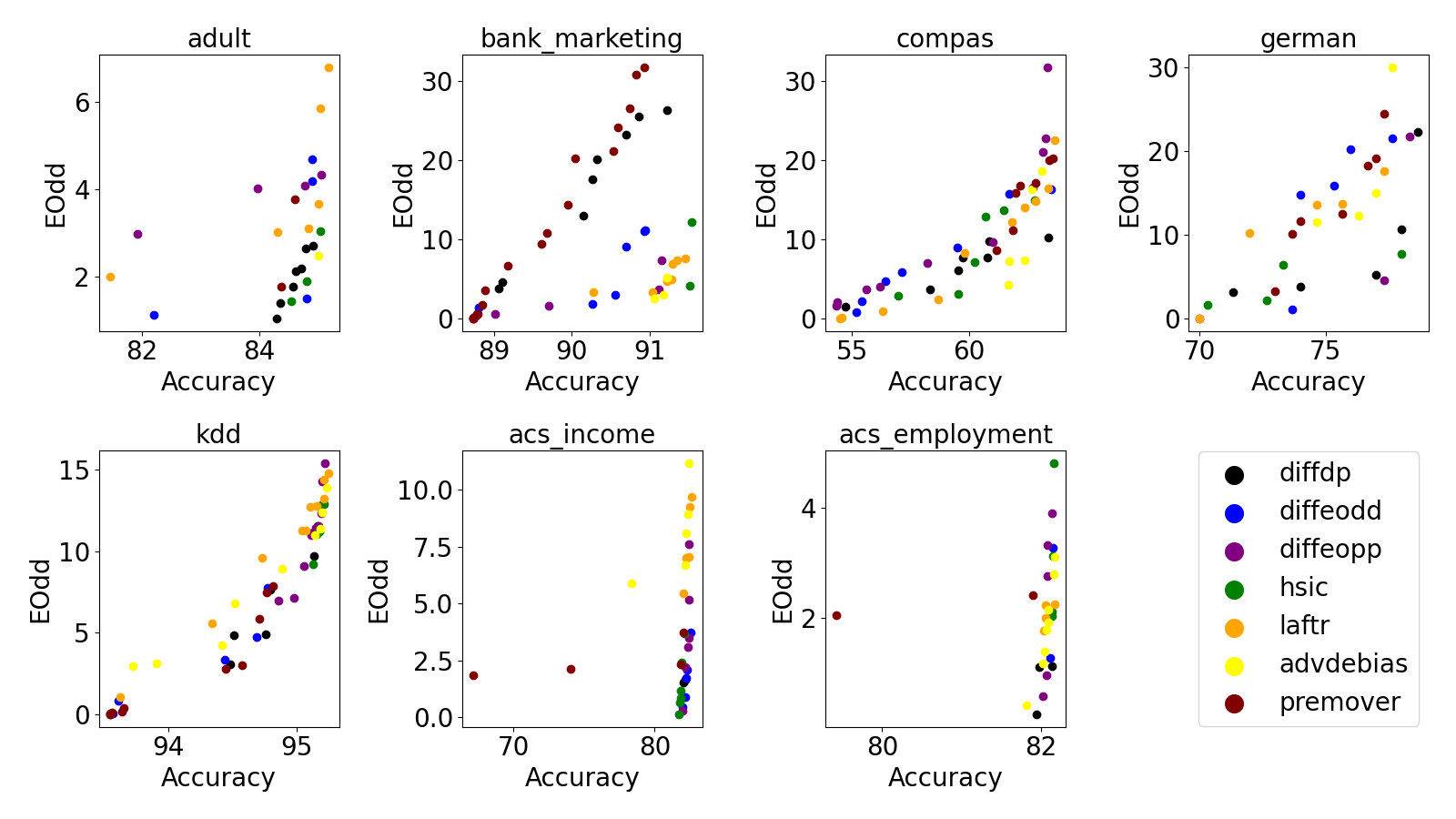}
    \caption{Pareto front of the fairness-utility (equalized odds-accuracy) tradeoff across various datasets. Each dot in the graph represents a separate training run on the pareto front with changing hyperparameters, random seeds and control parameters.}
    \label{fig:pareto_front_eodd}
\end{figure*}

\begin{figure*}
    \centering
    
    \includegraphics[width=0.98\linewidth]{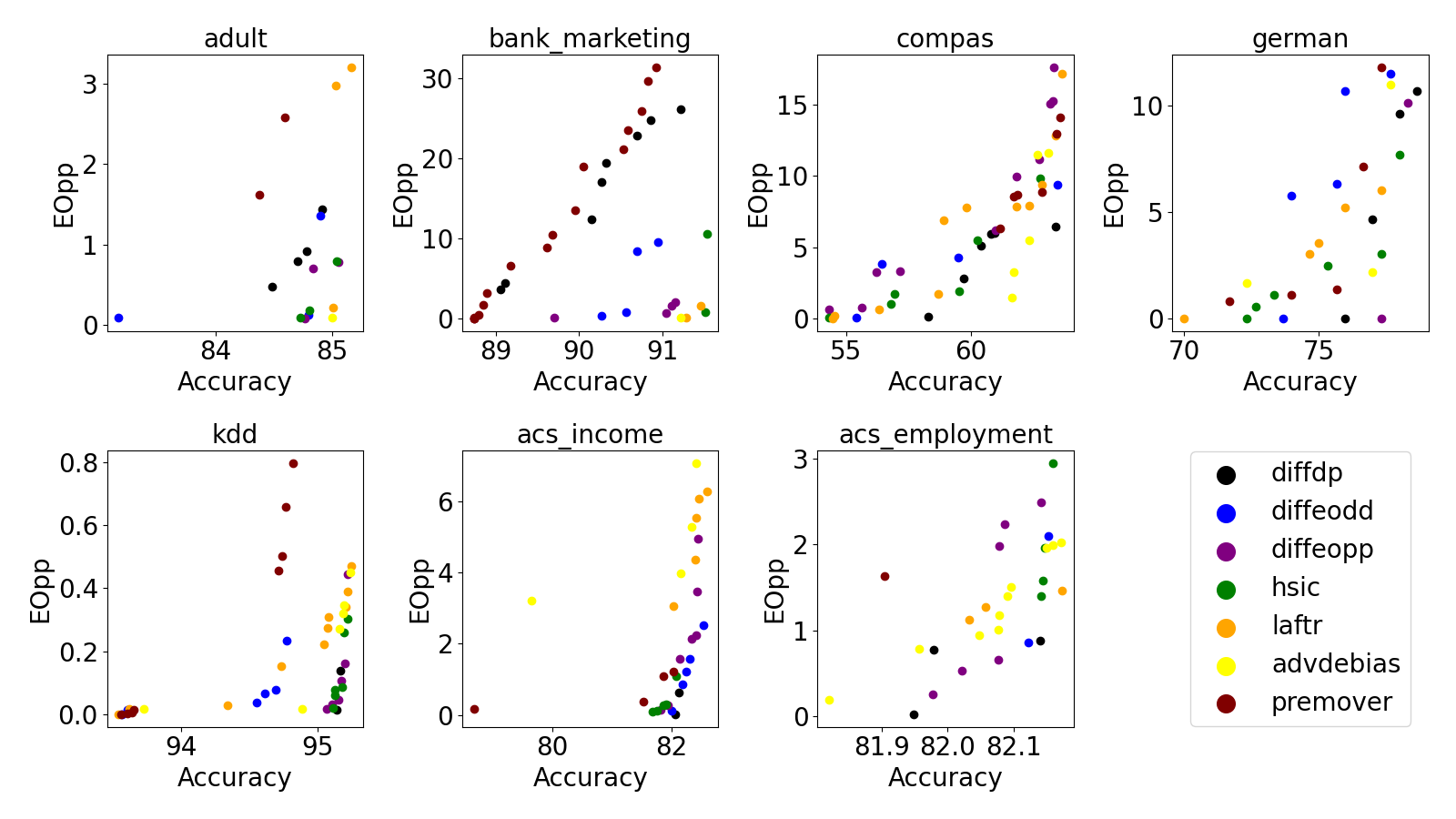}
    \caption{Pareto front of the fairness-utility (equal opportunity-accuracy) tradeoff across various datasets. Each dot in the graph represents a separate training run on the pareto front with changing hyperparameters, random seeds and control parameters.}
    \label{fig:pareto_front_eopp}
\end{figure*}

We present additional results on the pareto front for various algorithms and datasets, under fairness definition as equalized odds (Figure \ref{fig:pareto_front_eodd}) and equal opportunity (Figure \ref{fig:pareto_front_eopp}). Similar to the trends seen in the main paper, we find many different algorithms provide competitive tradeoffs when allowed to perform appropriate hyperparameter optimization.

\end{document}